\documentclass[preprint,12pt]{elsarticle}
\usepackage{amssymb}
\usepackage{amsmath}
\usepackage{amsthm}

\usepackage[total={6.5in, 9in}]{geometry}

\journal{Computer Methods in Applied Mechanics and Engineering}

\usepackage{cases}
\usepackage{algorithm}
\usepackage{algpseudocode}
\usepackage{url}
\usepackage{natbib}
\usepackage{xcolor}

\newtheorem*{remark}{Remark}

\newcommand{\bx}{\boldsymbol{x}}

\makeatletter
\DeclareRobustCommand{\cev}[1]{%
  \mathpalette\do@cev{#1}%
}
\newcommand{\do@cev}[2]{%
  \fix@cev{#1}{+}%
  \reflectbox{$\m@th#1\vec{\reflectbox{$\fix@cev{#1}{-}\m@th#1#2\fix@cev{#1}{+}$}}$}%
  \fix@cev{#1}{-}%
}
\newcommand{\fix@cev}[2]{%
  \ifx#1\displaystyle
    \mkern#23mu
  \else
    \ifx#1\textstyle
      \mkern#23mu
    \else
      \ifx#1\scriptstyle
        \mkern#22mu
      \else
        \mkern#22mu
      \fi
    \fi
  \fi
}

\usepackage{bm}
\usepackage{enumitem}
\usepackage{float}
\usepackage{subfig}

\newlength{\tempdima}
\newcommand{\rowname}[1]% #1 = text
{\rotatebox{90}{\makebox[\tempdima][c]{\textbf{#1}}}}

\renewcommand{\thesubfigure}{\alph{subfigure}}
\newcommand{\mycaption}[1]% #1 = caption
{\refstepcounter{subfigure}\textbf{(\thesubfigure) }{\ignorespaces #1}}

\newcommand{\f}{\frac}

\newcommand{\E}{ {\mathbb{E}} }

\newcommand{\be}{\begin{equation}}
\newcommand{\ee}{\end{equation}}

\begin{document}

\begin{frontmatter}



\title{An Ensemble Score Filter for Tracking High-Dimensional Nonlinear Dynamical Systems}


\author[label1]{Feng Bao} %% Author name
\ead{bao@math.fsu.edu}
\author[label2]{Zezhong Zhang} %% Author name
\ead{zhangz2@ornl.gov}
\author[label2]{Guannan Zhang\corref{cor1}} %% Author name
\ead{zhangg@ornl.gov}

\cortext[cor1]{Corresponding author}
\affiliation[label1]{organization={Department of Mathematics, Florida State University},%Department and Organization
            addressline={1017 Academic Way}, 
            city={Tallahassee},
            postcode={32306}, 
            state={FL},
            country={USA}}

\affiliation[label2]{organization={Computer Science and Mathematics Division, Oak Ridge National Laboratory},%Department and Organization
            addressline={1 Bethel Valley Rd}, 
            city={Oak Ridge},
            postcode={37831}, 
            state={TN},
            country={USA}}

%% Abstract
\begin{abstract}
We propose an ensemble score filter (EnSF) for solving high-dimensional nonlinear filtering problems with superior accuracy. A major drawback of existing filtering methods, e.g., particle filters or ensemble Kalman filters, is the low accuracy in handling high-dimensional and highly nonlinear problems. EnSF attacks this challenge by exploiting the score-based diffusion model, defined in a pseudo-temporal domain, to characterizing the evolution of the filtering density. EnSF stores the information of the recursively updated filtering density function in the score function, instead of storing the information in a set of finite Monte Carlo samples (used in particle filters and ensemble Kalman filters). Unlike existing diffusion models that train neural networks to approximate the score function, we develop a training-free score estimation that uses a mini-batch-based Monte Carlo estimator to directly approximate the score function at any pseudo-spatial-temporal location, which provides sufficient accuracy in solving high-dimensional nonlinear problems as well as saves a tremendous amount of time spent on training neural networks. High-dimensional Lorenz-96 systems are used to demonstrate the performance of our method. EnSF provides surprising performance, compared with the state-of-the-art Local
Ensemble Transform Kalman Filter method, in reliably and efficiently tracking extremely high-dimensional Lorenz systems (up to 1,000,000 dimensions) with highly nonlinear observation processes.
\end{abstract}


%% Keywords

\end{frontmatter}

\section{Introduction}\label{sec:intro}

Tracking high-dimensional nonlinear dynamical systems, also known as nonlinear filtering, represents a significant avenue of research in data assimilation, with applications in weather forecasting, material sciences, biology, and finance\cite{Bao_CiCP20, Bao_Cogan20, Multi-target_2007, Evense_EnKF, Filter_Finance}. 
The goal of addressing a filtering problem is to exploit noisy observational data streams to estimate the unobservable state of a stochastic dynamical system of interest. In linear filtering, where both the state and observation dynamics are linear, the Kalman filter provides an optimal estimate for the unobservable state, attainable analytically under the Gaussian assumption. Nevertheless, maintaining the covariance matrix of a Kalman filter is not computationally feasible for high-dimensional systems. For this reason, ensemble Kalman filters (EnKF) were developed in \cite{https://doi.org/10.1029/94JC00572,DataAssimilationUsinganEnsembleKalmanFilterTechnique,10.5555/1206873} to represent the distribution of the system state using a collection of state samples, called an ensemble, and to replace the covariance matrix in the Kalman filter with the sample covariance computed from the ensemble.
EnKF methods, especially the Local Ensemble Transform
Kalman Filter (LETKF) \cite{hunt_et_al_2007,10.1145/3581784.3627047}, are deployed operationally \cite{houtekamer_et_al_2014,schraff_et_al_2016} widely used to integrate observations for the purpose of understanding complex processes such as atmospheric convection \cite{aksoy_et_al_2009,aksoy_et_al_2010}. 
Despite many successful applications, EnKFs suffer from fundamental limitations as they make Gaussian assumptions in their update step, which leads to severe model bias in solving highly nonlinear systems. Hyper-parameters like localization and inflation have been added to EnKF to handle high-dimensional nonlinear problems. However, the accuracy of EnKF is extremely sensitive to the hyper-parameters, as demonstrated in Section \ref{sec:ex}, such that fine-tuning needs to be re-conducted whenever there is a small change, e.g., observation noise level, to the target filtering problem. Moreover, even though EnKF (e.g., LETKF\cite{hunt_et_al_2007,10.1145/3581784.3627047}) is known to have good scalability for CPU-based parallel computing platforms\cite{10.1145/3581784.3627047}, the existing EnKF algorithms are not suitable for modern GPU-based supercomputers because the parallelization of EnKF resulting from localization, i.e., decomposing the large covariance matrix into a large number of very small covariance matrices, cannot fully exploit the computing power of GPUs.

In addition to the EnKF, several effective methods have been developed to tackle nonlinearity in data assimilation. These methods include the particle filter \cite{particle-filter,doi:10.1073/pnas.0909196106,MCMC-PF, CT1, Kang-PF,  APF, Sny,doi:10.1137/20M1312204,Solonen_2016,10.2140/camcos.2010.5.221}, the Zakai filter \cite{Bao_Zakaid_2015, zakai}, and so on. For example, 
the particle filter employs a set of random samples, referred to as particles, to construct an empirical distribution to approximate the filtering density of the target state. Upon receiving observational data, the particle filter uses Bayesian inference to assign likelihood weights to the particles. A resampling process is iteratively performed, generating duplicates of particles with large weights and discarding particles with small weights. 
Although particle filters emerged around the same time as the EnKF, their implementation to large-scale models has been difficult due to the curse of dimensionality (weight collapse). This means that particle filters require prohibitively large ensemble sizes (number of particles) to ensure long-term stability. While there have been significant advances in this direction \citep{todter_et_al_2016,poterjoy_et_al_2017,rojahn_et_al_2023}, the resulting particle filers often provide marginal advantages over the state-of-the-art EnKFs used in operations.

In this work, we introduce a novel ensemble score filter (EnSF) that exploits the score-based diffusion model \cite{NEURIPS2021_49ad23d1,NEURIPS2020_4c5bcfec,NEURIPS2019_3001ef25,song2021scorebased} defined in a pseudo-temporal domain to characterize the evolution of the filtering density. The score-based diffusion model is a popular generative machine learning model for generating samples from a target distribution. Diffusion models have been applied to nonlinear filtering problems in our previous work \cite{bao2023scorebased}. Despite the promising performance, a major drawback of the method in \cite{bao2023scorebased} is that the neural network used to learn the score function needs to be re-trained at every filtering step after assimilating new observational data. Even though we can store the checkpoint of the neural network weights from previous filtering step, the neural network training still takes several minutes at each filtering step for a 100-dimensional Lorenz-96 model. Moreover, training the score function will require storing all the paths of the forward stochastic processes of the diffusion model, which leads to high storage requirements for high-dimensional problems. To resolve these issues, the key idea of EnSF is to completely avoid training neural networks to learn the score function. Instead, we derive the closed form of the score function and develop a 
training-free score estimation that uses mini-batch-based Monte Carlo estimators to directly approximate the score function at any pseudo-spatial-temporal location in the process of solving the reverse-time diffusion sampler. Numerical examples in Section \ref{sec:ex} demonstrate that the training-free score estimation approach can provide sufficient accuracy in solving high-dimensional nonlinear problems as well as save tremendous amount of time spent on training neural networks. Another essential aspect of EnSF is its analytical update step, which gradually incorporates data information into the score function. This step is crucial in mitigating the degeneracy issue faced when dealing with very high-dimensional nonlinear filtering problems. The main contributions of this work are summarized as follows:
\vspace{-0.0cm}
\begin{itemize}[leftmargin=20pt]\itemsep0.05cm
    \item We develop a training-free score function estimator that allows the diffusion model to be updated in real-time without re-training when new observational data is collected.  
    \item We showcase the remarkable robustness of the performance of EnSF with respect to its hyper-parameters by using a set of fine-tuned hyper-parameters to dramatically different scenarios, e.g., different dimensions, different observation noise, etc.   
    \item We showcase the superior performance of EnSF by comparing it with the state-of-the-art LETKF method in tracking 1,000,000-dimensional Lorenz-96 models with highly nonlinear observations. 
\end{itemize}

%
% The key idea of EnSF is to store the information of the filtering density in the score function, as opposed to storing the information in a set of finite random samples used in particle filters and ensemble Kalman filters. 
% Specifically, we propagate Monte Carlo samples through the state dynamics to generate ``data samples'' that follow the filtering density, and use those data samples to construct a score function. 
% Unlike existing diffusion models that train a neural network to approximate the score function\cite{song2021scorebased,bao2023scorebased,pmlr-v180-shi22a}, we 
% develop a training-free score estimation that uses mini-batch-based Monte Carlo estimators to directly approximate the score function at any pseudo-spatial-temporal location in the process of solving the reverse-time diffusion sampler. 

The rest of this paper is organized as follows. In Section \ref{prob:set}, we briefly introduce the nonlinear filtering problem. In Section \ref{sec:SNF}, we provide a comprehensive discussion to develop our EnSF method. In Section \ref{sec:ex}, we demonstrate the performance of the EnSF method in solving a Lorenz-96 tracking problem in high-dimensional space with highly nonlinear observation processes, and we shall also conduct a series of comparison studies between EnSF and the state-of-the-art LETKF method. Concluding remarks and future directions are given in Section \ref{sec:con}.

\section{Problem setting}\label{prob:set}
Nonlinear filtering is a process used to estimate the states of a system that evolves in time, where the system dynamics and the measurements are influenced by nonlinearities and noise. This estimation problem is complex because nonlinearity distorts the relationship between the observed measurements and the actual states, making linear prediction and update methods ineffective.
Specifically, we are interested in tracking the state of the following stochastic dynamical system:
\begin{equation}\label{NLF:state}
%\hspace{-2cm}
 X_{t} = f(X_{t-1}, \omega_{t-1}), 
\end{equation}
where $t$ is the discrete time index, $X_t \in \mathbb{R}^d$ is the state of the system governed by a physical model $f: \mathbb{R}^d \times \mathbb{R}^k \mapsto \mathbb{R}^d$, and $\omega_{t-1} \in \mathbb{R}^k$ is a random variable representing the uncertainty in $f$. This uncertainty may be caused by natural perturbations to the physical model, incomplete knowledge, or unknown features of the model $f$. The existence of such uncertainty makes direct estimation and prediction of the state of the dynamical model infeasible. To filter out the uncertainty and make accurate estimations of the state, we rely on partial noisy observations of the state $X_t$ given as follows:
\begin{equation}\label{NLF:observation}
 Y_{t} = g(X_{t})+ \varepsilon_{t},
\end{equation}
where $Y_{t} \in \mathbb{R}^r$ is the observational data collected by nonlinear observation function $g(X_{t})$, and the observation is also perturbed by a Gaussian noise $\varepsilon_{t} \sim \mathcal{N}(0, \Sigma_t)$.

The goal of the filtering problem is to find the best estimate, denoted by $\hat{X}_{t}$, of the hidden state $X_{t}$, given the observation information $\mathcal{Y}_{t}:= \sigma(Y_{1:t})$, which is the $\sigma$-algebra generated by the observational data up to the time instant $t$. In mathematics, such optimal estimate for $X_{t}$ is usually defined by the optimal filter, i.e., the conditional expectation for $X_{t}$ conditioning on the observational information:
% \begin{equation}\label{Expectation:S}
$\hat{X}_{t} := \E[X_{t} | \mathcal{Y}_{t}]$.
The standard approach for solving the optimal filtering problem is the Bayesian filter, in which we aim to approximate the conditional probability density function (PDF) of the state, denoted by $p_{X_{t} | Y_{1:t}}(x_{t} |{y}_{1:t})$, which is referred to as the filtering density. 
The general idea of the Bayesian filter is to recursively incorporate observational data to describe the evolution of the filtering density from time $t-1$ to $t$ in two steps, i.e., the prediction step and the update step.

In the \emph{prediction step}, we utilize the Chapman-Kolmogorov formula to propagate the state variable from $t-1$ to $t$  and obtain the \emph{prior filtering density}
\begin{equation}\label{Kolmogorov}
p_{X_{t}|Y_{1:t-1}}(x_{t}|{y}_{1:t-1}) = \int_{\mathbb{R}^d} p_{X_{t-1}| X_{t-1}}(x_{t} | x_{t-1})p_{X_t|Y_{1:t-1}}(x_{t-1}|{y}_{1:t-1}) dx_{t-1},
\end{equation}
where $p_{X_t|Y_{1:t-1}}(x_{t-1}|{y}_{1:t-1})$ is the posterior filtering density obtained at the time instant $t-1$, $p_{X_{t-1}| X_{t-1}}(x_{t} | x_{t-1})$ is the transition probability density derived from the state dynamics in Eq.~\eqref{NLF:state}, and $p_{X_{t}|Y_{1:t-1}}(x_{t}|{y}_{1:t-1})$ is the prior filtering density for the time instant $t$.

In the \emph{update step}, we combine the likelihood function with the prior filtering density to obtain the \emph{posterior filtering density}, i.e.,
\begin{equation}\label{Bayes}
p_{X_t|Y_{1:t}}(x_{t}|{y}_{1:t}) \propto {p_{X_t|Y_{1:t-1}}(x_{t}|{y}_{1:t-1}) \, p_{Y_{t}|X_t}(y_{t} | x_{t})},
\end{equation} 
where $p_{X_t|Y_{1:t-1}}(x_{t}|{y}_{1:t-1}) $ is the prior filtering density in Eq.~\eqref{Kolmogorov}, and the likelihood function $p_{Y_{t}|X_t}(y_{t} | x_{t})$ is defined by
\begin{equation}\label{Likelihood}
p_{Y_{t}|X_t}(y_{t} | x_{t}) \propto \exp\left[ -\f{1}{2} \big(g(x_{t}) - y_{t}\big)^{\top}\Sigma_t^{-1} \big(g(x_{t}) - y_{t} \big) \right],
\end{equation}
with $\Sigma_t$ being the covariance matrix of the random noise $\varepsilon_t$ in Eq.~\eqref{NLF:observation}. 
In this way, the filtering density is predicted and updated through formulas Eq.~\eqref{Kolmogorov} to Eq.~\eqref{Bayes} recursively in time. Note that both the prior and the posterior filtering densities in Eq.~\eqref{Kolmogorov} and Eq.~\eqref{Bayes} are defined as the continuum level, which is not practical. Thus, one important research direction in nonlinear filtering is to study how to accurately approximate the prior and the posterior filtering densities. 

In the next section, we introduce how to utilize score-based diffusion models to solve the nonlinear filtering problem. The diffusion model was introduced into nonlinear filtering in our previous work \cite{bao2023scorebased} in which the score function is approximated by training a deep neural network. Although the use of score functions provides accurate results, there are several drawbacks resulting from training neural networks. First, the neural network needs to be re-trained/updated at each filtering time step, which makes it computationally expensive. For example, it takes several minutes to train and update the neural-network-based score function at each filtering time step for solving a 100-D Lorenz-96 model \cite{bao2023scorebased}. Second, since neural network models are usually over-parameterized, a large number of samples are needed to form the training set to avoid over-fitting. Third, hyperparameter tuning and validation of the trained neural network introduces extra computational overhead. These challenges motivated us to develop the EnSF method that completely avoids neural network training in score estimation, in order to greatly expand the powerfulness of the score-based diffusion model in nonlinear filtering.

\section{The ensemble score filter (EnSF) method}\label{sec:SNF}
We now describe the details of the proposed EnSF method. Section \ref{EnSF_pred} introduces how to use the score-based diffusion model to store the information in the prior filtering density in the prediction step of nonlinear filtering. Section \ref{sec:update} recalls how to analytically incorporate the likelihood information to transform the prior score to the posterior score at the update step of nonlinear filtering. Section \ref{dis_EnSF} introduces the implementation details of EnSF and the discussion on its computational complexity.

\subsection{The prediction step of EnSF}\label{EnSF_pred}
The goal of the prediction step in EnSF is to develop a score-based diffusion model as a stochastic transport map between the prior filtering density $p_{X_{t}|{Y}_{1:t-1}}(x_t|y_{1:t-1})$ in Eq.~\eqref{Kolmogorov} and the standard normal distribution. To proceed, we first define a pseudo-temporal variable 
\begin{equation}\label{pseudo_time}
   \tau \in \mathcal{T} = [0,1],
\end{equation}
which is different from the temporal domain for defining the state and observation processes in Eq.~\eqref{NLF:state} and Eq.~\eqref{NLF:observation}. At the $t$-th filtering step, we can define the following forward SDE in the pseudo-temporal domain $\mathcal{T}$:
\begin{equation}\label{eq:forward}
Z_{t,\tau} = b(\tau) Z_{t,\tau} d\tau + \sigma(\tau) dW_\tau,
\end{equation}
where $W_\tau$ is a standard $d$-dimensional Brownian motion, $b(\tau)$ is the drift coefficient,  $\sigma(\tau)$ is the diffusion coefficient, and the subscript $(\cdot)_t$ indicates that the SDE is defined for the $t$-th filtering step. Note that Eq.~\eqref{eq:forward} is a \textit{linear} SDE. Thus its solution can be derived as 
\begin{equation}\label{SDE_solu}
    Z_{t,\tau} = Z_{t,0}\exp\left[\int_0^\tau b(s)ds\right] + \int_0^\tau\exp\left[\int_s^\tau b(r) dr\right]\sigma(s)dW_s.
\end{equation}

The forward SDE is used to transport any given initial random variable $Z_{t,0} \sim q_{Z_{t,0}}(z_{t,0})$ at $\tau = 0$ to the standard normal distribution $Z_{t,1} \sim \phi_{(0, \mathbf{I}_d)}(z_{t,1})$ at $\tau =1$, where $\phi_{(0, \mathbf{I}_d)}(\cdot)$ denotes the PDF of the standard normal distribution. 
There are many choices of $b(\tau)$ and $\sigma(\tau)$ given in the literature \cite{NEURIPS2020_4c5bcfec,song2021scorebased,10.1162/NECO_a_00142} to achieve the task. In this work, we use the following definition:
\begin{equation}\label{eq:cof}
\begin{aligned}
b(\tau) = \frac{{\rm d} \log \alpha_\tau}{{\rm d} \tau} \;\;\; \text{ and }\;\;\; \sigma^2(\tau) = \frac{{\rm d} \beta_\tau^2}{{\rm d}\tau} - 2 \frac{{\rm d}\log \alpha_\tau}{{\rm d}\tau} \beta_\tau^2,
\end{aligned}
\end{equation}
where the two processes $\alpha_\tau$ and $\beta_\tau$ are defined by
\begin{equation}\label{eq:ab}
\alpha_\tau = 1-\tau, \;\; \beta_\tau^2 = \tau \;\; \text{ for } \;\; \tau \in \mathcal{T}=[0,1].
\end{equation}
Substituting the definitions of $b(\tau)$ and $\sigma(\tau)$ into Eq.~\eqref{SDE_solu}, we can obtain that the conditional probability density function $q_{Z_{t,\tau}|Z_{t,0}}(z_{t,\tau}|z_{t,0})$ for any fixed value $z_{t,0}$ is the following Gaussian distribution: 
\begin{equation}\label{eq:gauss}
q_{Z_{t,\tau}|Z_{t,0}}(z_{t,\tau}|z_{t,0}) = \phi_{(\alpha_\tau z_{t,0}, \beta_\tau^2 \mathbf{I}_d)}(z_{t,\tau}|z_{t,0}), 
\end{equation}
where $\phi_{(\alpha_\tau z_{t,0}, \beta_\tau^2 \mathbf{I}_d)}(\cdot)$ is the standard normal PDF with mean $\alpha_\tau z_{t,0}$ and covariance matrix $\beta_\tau^2 \mathbf{I}_d$. 
The above equation immediately leads to 
\begin{equation}
    q_{Z_{t,1}|Z_{t,0}}(z_{t,1}|z_{t,0}) = q_{Z_{t,1}}(z_{t,1}) = \phi_{(0, \mathbf{I}_d)}(z_{t,1}),
\end{equation}
meaning that the forward SDE in Eq.~\eqref{eq:forward} can transport any initial distribution to the standard normal distribution at $\tau = 1$.

Let the initial state $Z_{t,0}$ of the forward SDE in Eq.~\eqref{eq:forward} follow the prior filtering density $p_{X_{t}|{Y}_{1:t-1}}(x_t|y_{1:t-1})$ in Eq.~\eqref{Kolmogorov}, we have
\begin{equation}\label{setting_prior}
    Z_{t,0} := X_t |{Y}_{1:t-1}\;\, \Longrightarrow \;\, q_{Z_{t,0}}(z_{t,0}) = p_{X_{t}|{Y}_{1:t-1}}(x_t|y_{1:t-1}),
\end{equation}
such that the forward SDE can transport the prior filtering density to the standard normal distribution. However, what we need is the transport model in the opposite direction, i.e., from $\tau=1$ to $\tau=0$. To do this, we construct the corresponding backward SDE 
\begin{equation}\label{DM:RSDE}
d{Z}_{t,\tau} = \left[ b(\tau){Z}_{t,\tau} - \sigma^2(\tau) S_{t|t-1}(Z_{t,\tau}, \tau)\right] d\tau + \sigma(\tau) d\cev{W}_\tau,
\end{equation}
where $\int \cdot d \cev{W}_\tau$ is a backward It\^o integral \cite{BDSDE}, $b(\tau)$ and $\sigma(\tau)$ are the same as in the forward SDE. The notation $S_{t|t-1}(Z_{t,\tau}, \tau)$ defines the score function associated with the diffusion model for the prior filtering density $p_{X_{t}|{Y}_{1:t-1}}$ in Eq.~\eqref{Kolmogorov}, i.e., 
\begin{equation}\label{prior_score}
    S_{t|t-1}(z_{t,\tau}, \tau) := \nabla_z \log q_{Z_{t,\tau}}({z}_{t,\tau}),
\end{equation}
where the subscript $(\cdot)_{t|t-1}$ indicates that it is the prior score function without assimilating the observational data at the $t$-th filtering step. 

The methodology of EnSF is established based on the following derivation.  First, we observe that the probability density function $q_{Z_{t,\tau}}$ can be expressed as follows:
\[
q_{Z_{t,\tau}}(z_{t,\tau})= \int_{\mathbb{R}^d} q_{Z_{t,\tau},Z_{t,0}}(z_{t,\tau}, z_{t,0}) dz_{t,0} = \int_{\mathbb{R}^d}  q_{{Z}_{t,\tau} | Z_{t,0}}({z}_{t,\tau} | z_{t,0}) q_{Z_{t,0}}(
z_{t,0}) dz_{t,0}
\]
Then, by substituting this equation into Eq.~\eqref{prior_score} and exploiting the fact in Eq.~\eqref{eq:gauss}, we can rewrite the score function in the form of the following integral:
\begin{equation}\label{eq:score11}
\begin{aligned}
& S_{t|t-1}(z_{t,\tau}, \tau)\\
  =\, & \nabla_z \log \left(\int_{\mathbb{R}^d} q_{{Z}_{t,\tau} | Z_{t,0}}({z}_{t,\tau} | z_{t,0}) q_{Z_{t,0}}(
z_{t,0}) dz_{t,0}\right)\\
 =\, & \frac{1}{\int_{\mathbb{R}^d} q_{{Z}_{t,\tau} | Z_{t,0}}({z}_{t,\tau} | z'_{t,0}) q_{Z_{t,0}}(
z'_{t,0}) dz'_{t,0}}   \int_{\mathbb{R}^d}  - \frac{z_{t,\tau} - \alpha_\tau z_{t,0}}{\beta^2_\tau}q_{Z_{t,\tau}|Z_{t,0}}({z}_{t,\tau} | z_{t,0}) q_{Z_{t,0}}(
z_{t,0}) dz_{t,0}\\[5pt]
=\, &   \int_{\mathbb{R}^d}  - \frac{z_{t,\tau}- \alpha_\tau z_{t,0}}{\beta^2_\tau} w({z}_{t,\tau},  z_{t,0})  q_{Z_{t,0}}(z_{t,0})dz_{t,0},\\
\end{aligned}
\end{equation}
where the weight function $w({z}_{t,\tau},  z_{t,0})$ is defined by
\begin{equation}\label{eq:weight}
w({z}_{t,\tau},  z_{t,0}) := \frac{ q_{Z_{t,\tau}|Z_{t,0}}({z}_{t,\tau} | z_{t,0}) }{\int_{\mathbb{R}^d} q_{Z_{t,\tau}|Z_{t,0}}({z}_{t,\tau} | z'_{t,0})  q_{Z_{t,0}}(
z'_{t,0}) dz'_{t,0}},
\end{equation}
satisfying that $\int_{\mathbb{R}^d}w({z}_{t,\tau},  z_{t,0}) q_{Z_{t,0}}(z_{t,0})dz_{t,0} = 1$. 

Thus, the backward SDE and the score function fully characterize the prior filtering density in Eq.~\eqref{Kolmogorov}. When applying proper numerical schemes to approximate the score function $S_{t|t-1}(z_{t,\tau}, \tau)$ in Eq.~\eqref{eq:score11} and the backward SDE in Eq.~\eqref{DM:RSDE}, we can generate an unlimited number of samples from the prior filtering density. The implementation detail of the prediction step is given in Section \ref{dis_EnSF}. 

\subsection{The update step of EnSF}\label{sec:update}
The goal of the update step in EnSF is to incorporate the new observational data $Y_t$, or more specifically the likelihood function in Eq.~\eqref{Likelihood}, obtained at the $t$-th filtering step to update the prior filtering density in a Bayesian fashion. In the context of diffusion models, this task becomes how to update the prior score function $S_{t|t-1}$ to the posterior score function, denoted by $S_{t|t}$, by incorporating the likelihood function. In this work, we use a similar strategy as our previous work in \cite{bao2023scorebased} to analytically add the likelihood information to the prior score function $S_{t|t-1}$ to obtain a posterior score function $S_{t|t}$. 

Specifically, we assume the posterior score function has the following structure:
\begin{equation}\label{eq:d}
{S}_{t|t}(z_{t,\tau}, \tau) := {S}_{t|t-1}(z_{t,\tau}, \tau) + h(\tau) \nabla_z \log p_{Y_{t}|X_t}(y_{t} | z_{t,\tau}),
\end{equation}
where ${S}_{t|t-1}(z_{t,\tau}, \tau)$ is the prior score function in Eq.~\eqref{eq:score11}, and $\nabla_z \log p_{Y_{t}|X_t}(y_{t} | z_{t,\tau})$ is the gradient of the log likelihood function evaluated at $z_{t,\tau}$. According to practical usage of nonlinear filtering in numerical weather forecasting (e.g., \cite{IFS}), the analytical formula of the observation operator, i.e., the function $g$ in Eq.~\eqref{NLF:observation} is usually known, such that it is reasonable to assume the gradient of the log-likelihood is accessible in EnSF. The key component in the update step is the   
damping function $h(\tau)$ satisfying
\begin{equation}\label{eq:ht}
h(\tau) \text{ is monotonically decreasing in } [0,1] \text{ with } h(1) = 0 \text{ and } h(0) = 1,
\end{equation}
which determines how the likelihood information is gradually introduced into the score function while solving the backward SDE. In this work, we use $h(\tau) = 1-\tau$ in the numerical experiments. The likelihood has almost no influence on the prior score when the pseudo time $\tau$ is close to 1. As $\tau$ decreases, the diffusion term becomes less dominating and the likelihood information is gradually injected into the backward SDE via the drift term.

We emphasize that even though the proposed structure of the posterior score works very well for the numerical examples in Section \ref{sec:ex}, there may be a model structure error in the proposed posterior score function in Eq.~\eqref{eq:d} depending on the choice of $h(\tau)$. In other words, the proposed $S_{t|t}$ in Eq.~\eqref{eq:d} may not be the score associated with the exact posterior filtering density in 
Eq.~\eqref{Bayes}. Correcting the model error will ensure the theoretical rigor but may significantly increase the computational cost, especially for nonlinear filtering in which the score function needs to be updated dynamically in real time. 
Thus, how to develop a computationally efficient model error correction scheme is still an open question and will be considered in our future work. 

\begin{remark}[Avoiding the curse of dimensionality]
Incorporating the analytical form of the likelihood information, i.e., $\nabla_z \log p_{Y_{t}|X_t}(y_{t} | z_{t,\tau})$, into the score function plays a critical role in avoiding performing high-dimensional approximation, i.e., the curse of dimensionality, in the update step. In other words, when $\nabla_z \log p_{Y_{t}|X_t}(y_{t} | z_{t,\tau})$ is given, either in the analytical form or via automatic differentiation, we do not need to perform any approximation in $\mathbb{R}^d$. In comparison, EnKF requires approximating the covariance matrix and the particle filter requires construction of empirical distributions, both of which involve approximation of the posterior distribution in $\mathbb{R}^d$.  
\end{remark}

\subsection{Implementation of EnSF}\label{dis_EnSF}
We focus on how to discretize EnSF and approximate the score functions $S_{t|t}$ and $S_{t+1|t}$ in order to establish a practical implementation for EnSF. The classic diffusion model methods \cite{NEURIPS2020_4c5bcfec,song2021scorebased,10.1162/NECO_a_00142} train neural networks to learn the score functions. This approach works well for static problems that does not require fast evolution of the score function. However, this strategy becomes inefficient in solving the nonlinear filtering problem \cite{bao2023scorebased}, especially for extremely high-dimensional problems. To address this challenge, we propose a training-free score estimation approach that uses the Monte Carlo method to directly approximate the expression of the score function in Eq.~\eqref{eq:score11}, which enables extremely efficient implementation of EnSF. 

\subsubsection{Introducing two hyper-parameters into EnSF}\label{sec:hyper}
The advantage of the choice of the drift and diffusion coefficients in Eq.~\eqref{eq:cof} and Eq.~\eqref{eq:ab} is that the resulting forward SDE can map any distribution to the standard normal distribution within a bounded pseudo time interval $\mathcal{T} = [0,1]$. However, this approach also introduces several computational issues into the diffusion model. The first one is that the denominator in Eq.~\eqref{eq:score11} goes to zero as $\tau \rightarrow 0$, which will cause the explosion of the score function when $\beta_\tau^2 = \tau$ at $\tau = 0$; the second one is that when $\alpha_\tau = 1-\tau$ and the backward SDE is solved exactly, the conditional distribution in Eq.~\eqref{eq:gauss} indicates that 
the backward SDE will drive each path of the state $Z_{t,\tau}$ of the diffusion model to the infinitesimal neighborhood of one of the samples of $Z_{t,0}$ as $\tau \rightarrow 0$, which will limit the exploration power of the diffusion model.
To regularize the backward SDE, we introduce two hyperparameters, denoted by $\epsilon_\alpha$ and $\epsilon_\beta$, to the definitions of $\alpha_\tau$ and $\beta_\tau$ in Eq.~\eqref{eq:ab}, respectively. After re-parameterization, the actual $\alpha_\tau$ and $\beta_\tau$ used in our EnSF implementation are
\begin{equation}\label{eq:hyper}
    \bar{\alpha}_{\tau} = 1 - \tau(1 - \epsilon_{\alpha}); \quad\quad \bar{\beta}^2_{\tau} = \epsilon_{\beta} + \tau(1 - \epsilon_{\beta}).
\end{equation}
Based on the above equation, $\bar{\alpha}_{\tau}$ is a linear interpolation between $(0,1)$ and $(1, \epsilon_{\alpha})$, and $\bar{\beta}^2_{\tau}$ is a linear interpolation between $(0, \epsilon_{\beta})$ and $(1,1)$. 
The fine-tuning procedure shown in Section \ref{sec:compare} indicates that even though the performance of EnSF is not sensitive to the two hyper-parameters compared to LETKF, the fine-tuning can still provide significant performance improvement.

\subsubsection{Training-free score estimation}\label{sec:train}
Unlike existing methods that use neural network models to learn the score function, in this work we propose to directly discretize the score representation in Eq.~\eqref{eq:score11}. Specifically, we assume that we are given a set of samples $\{x_{t-1,j}\}_{j=1}^J$ drawn from the posterior filtering density function $p_{X_{t-1,\tau}|Y_{1:t-1}}(x_{t-1,\tau} | y_{1:t-1})$ from previous filtering time step $t-1$. For any fixed pseudo-time instant $\tau \in \mathcal{T}$ and $z_{t,\tau} \in \mathbb{R}^d$, the integral in Eq.~\eqref{eq:score11} can be estimated by 
\begin{equation}\label{eq:MC}
S_{t|t-1}(z_{t,\tau}, \tau) \approx \bar{S}_{t|t-1}(_{t,\tau}, \tau) :=  \sum_{n=1}^{N} - \frac{z_{t,\tau} - \bar{\alpha}_\tau f(x_{t-1,j_n},\omega_{t-1,j_n})}{\bar{\beta}^2_\tau} \bar{w}({z},  f(x_{t-1,j_n},\omega_{t-1,j_n})), 
\end{equation}
using a mini-batch $\{x_{t-1,j_n}\}_{n=1}^N$ with batch size $N \le J$, where $f(\cdot,\cdot)$ is the state equation in Eq.~\eqref{NLF:state}. The weight $w({z}_{t,\tau}, f(x_{t-1,j_n},\omega_{t-1,j_n}))$ in Eq.~\eqref{eq:weight} is approximated by
\begin{equation}\label{eq:weight_app}
 \bar{w}({z}_{t,\tau},   f(x_{t-1
,j_n},\omega_{t-1,j_n})) := \frac{q_{Z_{t,\tau}|Z_{t,0}}(z_{t,\tau} |  f(x_{t-1,j_n},\omega_{t-1,j_n})) }{\sum_{m=1}^{N} q_{Z_{t,\tau}|Z_{t,0}}(z_{t,\tau}|  f(x_{t-1,j_{m}},\omega_{t-1,j_m}))},
\end{equation}
which means $w({z}_{t,\tau},  f(x_{t-1,j_n},\omega_{t-1,j_n}))$ can be estimated by the normalized probability density values $\{q_{Z_{t,\tau}|Z_{t,0}}(z_{t,\tau}|f(x_{t-1,j_n},\omega_{t-1,j_n}))\}_{n=1}^N$. In practice, the mini-batch $\{x_{t-1,j_n}\}_{n=1}^N$ could be a very small subset of $\{x_{t-1,j}\}_{j=1}^J$ to ensure sufficient accuracy in solving the filtering problems. In fact, we use batch size one for our mini-batch in the numerical experiments in Section \ref{sec:ex}, which provides satisfactory performance. The training-free score estimation is significantly more efficient than training neural networks to learn the score function \cite{bao2023scorebased} in nonlinear filtering where the posterior filtering density needs to be updated frequently.

\subsubsection{Summary of EnSF workflow}\label{app:discrete}
Now we combine the aforementioned approximation schemes to develop a detailed algorithm to evolve the filtering density from $p_{X_{t-1}|Y_{1:t-1}}(x_{t-1}|{y}_{1:t-1})$ to $p_{X_{t}|Y_{1:t}}(x_{t}|{y}_{1:t})$. At the $t$-th filtering step, we assume we are given a set of samples $\{x_{t-1,j}\}_{j=1}^J$ drawn from the posterior filtering density function $p_{X_{t-1}|Y_{1:t-1}}(x_{t-1}|{y}_{1:t-1})$ and the goal is to generate a set of samples $\{x_{t,j}\}_{j=1}^J$ from $p_{X_{t}|Y_{1:t}}(x_{t}|{y}_{1:t})$. This evolution involves the simulation of the backward SDE of the diffusion model driven by the approximate score $S_{t|t}$. {Even though the forward SDE is included in the diffusion model, the training-free score estimation approach allows us to skip the simulation of the forward SDE.}.   

We use the Euler-Maruyama scheme to discretize the backward SDE. Specifically, we introduce a partition of the pseudo-temporal domain $\mathcal{T} = [0,1]$, i.e.,
\[
\mathcal{D}_K : = \{\tau_k \;| \;0 = \tau_0 < \tau_1  < \cdots < \tau_k < \tau_{k+1} < \cdots < \tau_K = 1\}
\] 
with uniform step-size $\Delta \tau = {1}/{K}$. We first draw a set of samples $\{z_{t,1}^j\}_{j=1}^J$ from the standard normal distribution. For each sample $z_{t,1}^j$, we obtain the approximate solution $z_{t,0}^j$ by recursively evaluating the following scheme
\begin{equation}\label{Euler:RSDE}
\begin{aligned}
z_{t,\tau_k}^j = z_{t,\tau_{k+1}}^j & - \big[ b(\tau_{k+1} ) z_{t,\tau_{k+1}}^j   - \sigma^2(\tau_{k+1}) \bar{S}_{t|t}(z_{t,\tau_{k+1}}^j, \tau_{k+1}) \big] \Delta \tau + \sigma(\tau_{k+1}) \Delta W_{\tau_{k+1}}^j, 
\end{aligned}
\end{equation}
for $k = K-1, K-2, \cdots, 1,0$, where $\Delta W_{\tau_{k+1}}^j$ is a realization of the Brownian increment, and the approximate score function is calculated by
\begin{equation}\label{eq:score_update}
    \bar{S}_{t|t}(z_{t,\tau_{k+1}}^j, \tau_{k+1}) = 
\bar{S}_{t|t-1}(z_{t,\tau_{k+1}}^j, \tau_{k+1}) + h(\tau_{k+1}) \nabla_z \log p_{Y_{t}|X_t}(y_{t} | z_{t,\tau_{k+1}}^j).
\end{equation}
Since the backward SDE is driven by $\bar{S}_{t|t}$, we treat $\{z_{t,0}^j\}_{j=1}^J$ as the desired sample set $\{x_{t,j}\}_{j=1}^J$ from the posterior filtering density $p_{X_{t}|Y_{1:t}}(x_{t}|{y}_{1:t})$. EnSF workflow is summarized as a pseudo-algorithm in Algorithm 1. 

\noindent\makebox[\linewidth]{\rule{\textwidth}{0.5pt}}\\
\vspace{-0.5cm}
\newline {\bf Algorithm 1: the pseudo-algorithm for EnSF}\vspace{-0.2cm} \\ 
\noindent\makebox[\linewidth]{\rule{\textwidth}{0.5pt}}
\vspace{-0.3cm}
\newline1:\, {\bf Input}: the state equation $f(X_t, \omega_t)$, the prior density $p_{X_0}(x_0)$;
\vspace{0.1cm}
\newline2: Generate $J$ samples $\{x_{0,j}\}_{j=1}^J$ from the prior $p_{X_0}(x_0)$;
\vspace{0.1cm}
\newline3: \,{\bf for} $t = 1, \ldots, $
\vspace{0.1cm}
\newline4: \qquad Run the state equation in Eq.~\eqref{NLF:state} to get predictions $\{f(x_{t-1,j},\omega_{t-1,j})\}_{j=1}^J$;
\vspace{0.1cm}
\newline5: \qquad {\bf for} $k = K-1, \ldots, 0 $
\vspace{0.1cm}
\newline6: \qquad \quad\; Compute the weight $\{ \{\bar{w}({z}_{t,\tau}^j,f(x_{t-1,j_n},\omega_{t-1,j_n}))\}_{n=1}^N\}_{j=1}^J$ using Eq.~\eqref{eq:weight_app};
\vspace{0.1cm}
\newline7: \qquad \quad\; Compute $\{\bar{S}_{t|t-1}(z_{t,\tau_{k+1}}^j, \tau_{k+1})\}_{j=1}^J$ using Eq.~\eqref{eq:MC};
\vspace{0.1cm}
\newline8: \qquad \quad\; Compute $\{\bar{S}_{t|t}(z_{t,\tau_{k+1}}^j, \tau_{k+1})\}_{j=1}^J$ using Eq.~\eqref{eq:score_update};
\vspace{0.1cm}
\newline9: \qquad \quad\; Compute $\{z_{t,\tau_k}^j \}_{j=1}^J$ using Eq.~\eqref{Euler:RSDE};
\vspace{0.1cm}
\newline11: \quad\; {\bf end}\vspace{0.1cm} 
\newline10: \quad\; Let $\{x_{t,j}\}_{j=1}^J = \{z_{t,0}^j\}_{j=1}^J $; 
\vspace{0.1cm}
\newline11: {\bf end}\vspace{-0.1cm} \\
\noindent\makebox[\linewidth]{\rule{\textwidth}{0.5pt}}

 \subsubsection{Discussion on the computational complexity of EnSF}\label{sec:comp}
Since the cost of running the state equation $f(X_t,\omega_t)$ in Eq.~\eqref{NLF:state} is problem-dependent, we only discuss the cost of the matrix operations for Line 6 -- 9 in Algorithm 1. 
In terms of the storage cost, the major storage of EnSF is used to store the two sample sets, i.e., $\{x_{t,j}\}_{j=1}^J$ from the posterior filtering density of the previous time step and $\{z_{\tau,j}\}_{j=1}^J$ for the states of the diffusion model. Each set is stored as a matrix of size $J \times d$ where $J$ is the number of samples and $d$ is the dimension of the filtering problem. The storage requirement is suitable for conducting all the computations on modern GPUs. In terms of the number of floating point operations, Line 6 -- 9 in Algorithm 1 for fixed $t$ and $\tau$ involves $\mathcal{O}({J\times N \times d})$ operations including element-wise operations and matrix summations, where $N< J$ is the size of the mini-batch used to estimate the weights in Eq.~\eqref{eq:weight_app}. So the total number of floating point operations is on the order of $\mathcal{O}({J\times N \times d \times K})$ to update the filtering density from $t$ to $t+1$, where $K$ is the number of time steps for discretizing the backward SDE. 
The numerical experiments in Section \ref{sec:ex} show that 
the number of samples $J$ can grow very slowly with the dimension $d$ while maintaining a satisfactory performance for tracking the Lorenz-96 model, which indicates the superior efficiency of EnSF in handling extremely high-dimensional filtering problems.

\section{Numerical experiments: tracking the 1,000,000-dimensional Lorenz-96 model}\label{sec:ex}
We demonstrate EnSF's capability in handling a high-dimensional Lorenz-96 model. 
Specifically, we track the state of the Lorenz-96 model described as follows:
\begin{equation}\label{Lorenz}
\f{dx_i}{dt} = (x_{i+1} - x_{i-2}) x_{i-1} + F, \qquad i = 1, 2, \cdots, d, \quad d \geq 4,
\end{equation}
where $X_t = [x_1(t), x_2(t), \cdots, x_d(t)]^{\top}$ is a $d$-dimensional target state, and it is assumed that $x_{-1} = x_{d-1}$, $x_{0} = x_{d}$, and $x_{d+1} = x_1$. The term $F$ is a forcing constant. When $F=8$, the Lorenz-96 dynamics \eqref{Lorenz} becomes a chaotic system, which makes tracking the state $X_t$ a challenging task for all the existing filtering techniques, especially in high dimensional spaces. In our numerical experiments, we discretize Eq.~\eqref{Lorenz} through the Runge-Kutta (RK4) scheme. To avoid NaN values in the experiments, we clip the solutions of the forward solver at a magnitude of 50. Specifically, we set the ensemble values to 50 or -50 when they exceed this range. To initialize the Lorenz-96 system, we first pick a random sample from $ N(\mathbf{0}, 3^2 \mathbf{I}_{d})$ and then run $1000$ burn-in simulation steps through the RK4 scheme to obtain our true initial state $X_0$. Our initial guess for the initial ensemble $X_0$ is a standard Gaussian random variable $N(\mathbf{0}, \mathbf{I}_{d})$, which means that we do not possess any effective information about $X_0$ at the beginning.

Since EnSF is designed as a nonlinear filter for high-dimensional problems, we carry out experiments on one million-dimensional Lorenz-96 system, i.e., $d = 1,000,000$, where the observational process in Eq.~\eqref{NLF:observation} is an arctangent function of the state, i.e.,
\begin{equation}\label{Measurement}
Y_{t+1} = \arctan( X_{t+1} ) + \varepsilon_{t+1}.
\end{equation}
The chaotic state dynamics in Eq.~\eqref{Lorenz} along with the highly linear observation in Eq.~\eqref{Measurement} would make the tracking of the Lorenz-96 system extremely challenging, especially in such high-dimensional space. 
In what follows, we shall demonstrate the performance of EnSF in solving the above Lorenz-96 tracking problems in one-million-dimensional space, and we shall also carry out a series of experiments to compare our method with the state-of-the-art optimal filtering method, i.e., the Local Ensemble Transform Kalman Filter (LETKF), which is the method adopted by the European Center for Medium-Range Weather Forecasts for hurricane forecasting. 

\begin{remark}[{\bf Reproducibility}]
EnSF method for the high-dimensional Lorenz-96 problem is implemented in Pytorch with GPU. The source code is publicly available at \url{https://github.com/zezhongzhang/EnSF}. The numerical results in this section can be exactly reproduced using the code on Github.
\end{remark}

\subsection{Illustration of EnSF's accuracy}\label{sec:baseline}
In the first experiment, we illustrate the accuracy of EnSF in tracking the 1,000,000-dimensional Lorenz-96 model, and we track the target state $800$ time steps with temporal step size $\Delta t = 0.01$ and observational noise $\varepsilon_t \sim(\mathbf{0}, 0.05^2\mathbf{I}_{d})$.

\begin{figure}[h!]
\centering
\includegraphics[width=0.75\textwidth]{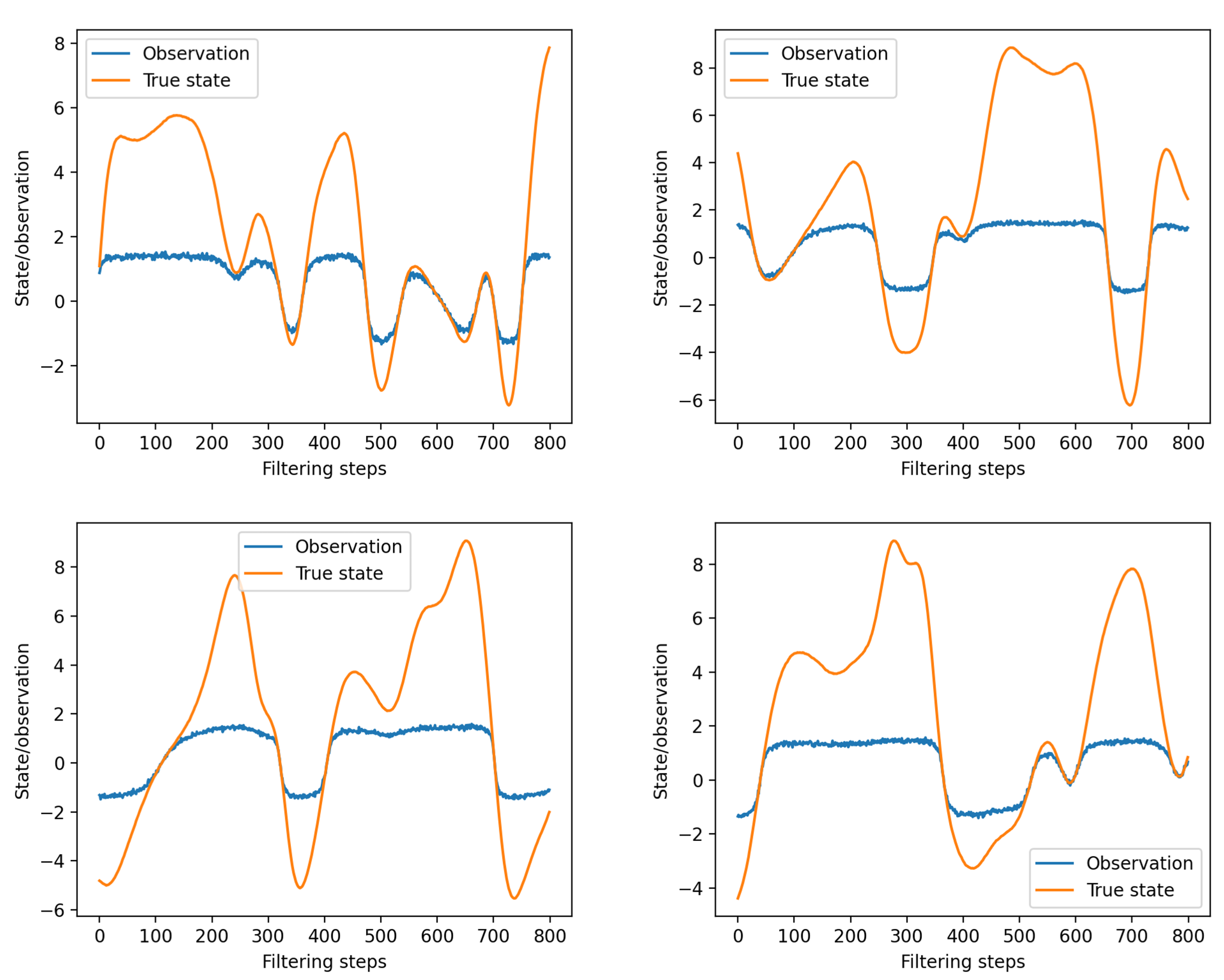}
\caption{Illustration of the nonlinearity of observation process by comparing the true state $X_{t}$ and the observation $Y_t$ along four randomly selected directions. Due to the nonlinearity of $arctan()$, the observation $Y_t$ does not provide sufficient information of the state when $X_t$ is outside the domain $[-\pi/2,\pi/2]$.}\label{obsshow}
% \vspace{-0.4cm}
\end{figure}

In Figure \ref{obsshow}, we illustrate the nonlinearity of the observation system by comparing the true state $X_{t}$ and the observation $Y_t$ along four randomly selected directions. Due to the nonlinearity of $arctan()$, the observation $Y_t$ does not provide sufficient information of the state when $X_t$ is outside the domain $[-\pi/2,\pi/2]$. When it happens, the partial derivative of $Y_t$ is very close to zero such that there is very little update of the score function in Eq.~\eqref{eq:d} along the directions with states outside $[-\pi/2,\pi/2]$. In other words, there may be only a small subset of informative observations at each filtering time step.
\begin{figure}[h!]
\centering
\includegraphics[width=0.75\textwidth]{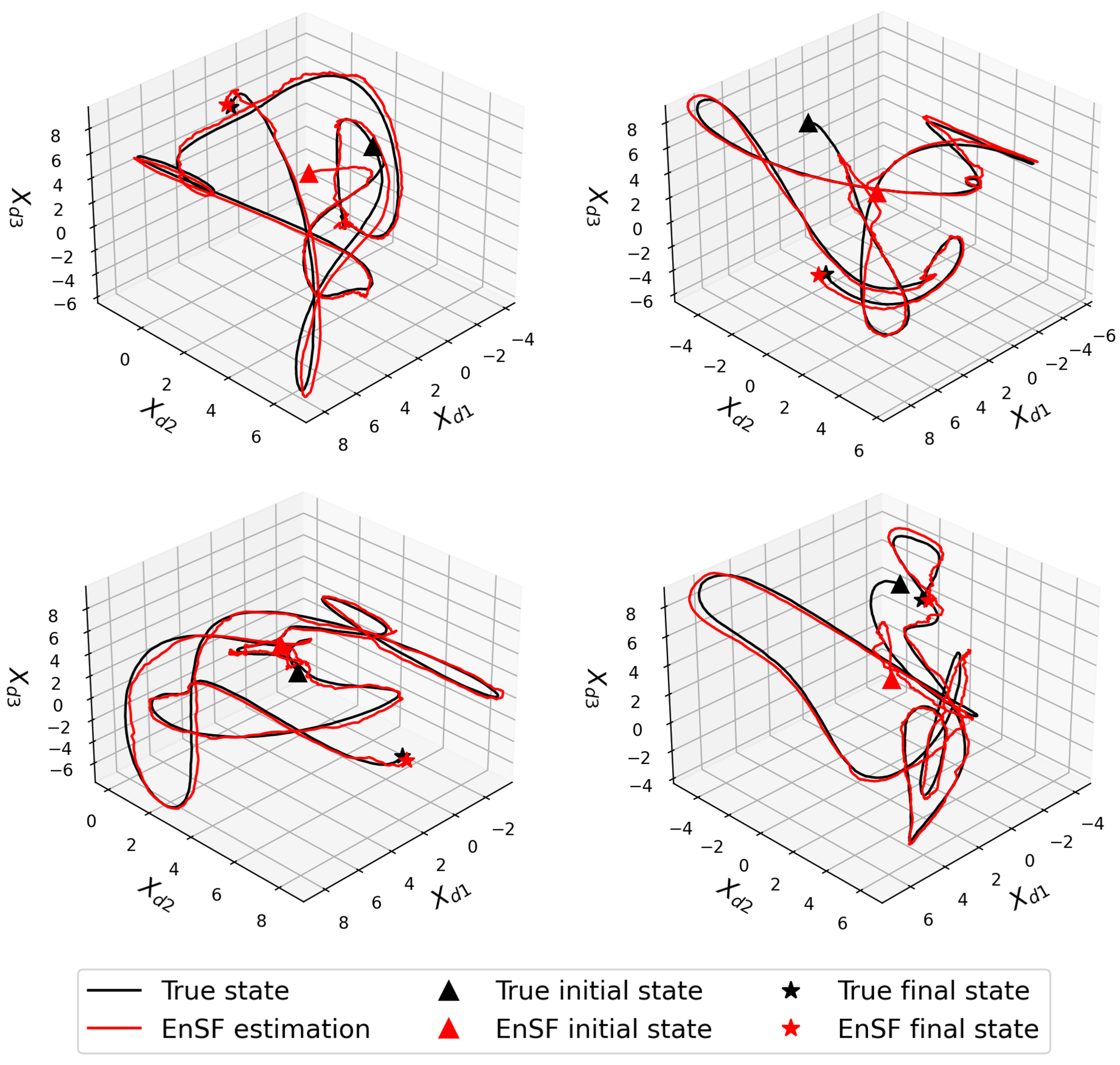}
\caption{Comparison between the true state trajectories and the estimated trajectories obtained by EnSF, each sub-figure shows the trajectories along randomly selected three directions in the $1,000,000$-dimensional state space. We observe that even though the initial guess for EnSF is far from the true initial state, EnSF gradually captures the true state by assimilating the observational data after several filtering steps, providing sufficient accuracy in capturing such a high-dimensional chaotic system.}\label{path}
% \vspace{-0.4cm}
\end{figure}

Figure \ref{path} shows the comparison between the true state trajectories and the estimated trajectories, each sub-figure shows the trajectories along randomly selected three directions in the 1,000,000-dimensional state space. EnSF is implemented with $500$ pseudo time steps when discretizing the backward SDE, and the ensemble size that we picked is $20$ samples. Since EnSF's initial estimate is randomly sampled from $\mathcal{N}(\mathbf{0}, \mathbf{I}_d)$, it is far from the true initial state. After several filtering steps, EnSF gradually captures the true state by assimilating the observational data. Even though there are some discrepancy between the true and the estimated states, the accuracy of EnSF is sufficient for capturing such a high-dimensional chaotic system.

% Figure \ref{rmse_all} shows root mean square errors (RMSEs) of EnSF's state estimation using $20$ repeated trials along with its $90\%$ confidence band. The RMSEs are averaged over the $1,000,000$ dimensional state space and $20$ trials. The initial error is relatively big due to the discrepancy between EnSF's initial guess and the true state and the error gradually reduces as observational data are assimilated.

% %
% \begin{figure}[h!]
% \centering
% \includegraphics[width=0.8\textwidth]{./figs/RMSE_all.png}
% \caption{RMSE of EnSF's state estimation using 20 repeated trials with different initializations.The RMSE is averaged over the state space and 90\% confidence band is computed using the 20 trials.}\label{rmse_all}
% \end{figure}
%

\subsection{Comparison between EnSF and  LETKF}\label{sec:compare}
In the following numerical experiments, we compare EnSF with LETKF in tracking the $1,000,000$-dimensional Lorenz-96 model, and we track the target state $1500$ time steps with temporal step size $\Delta t = 0.01$ and observational noise $\varepsilon_t \sim(\mathbf{0}, 0.05^2\mathbf{I}_{d})$. To allow gaps between prediction and update, we implement the Bayesian update procedure with time step size $0.1$, i.e., we implement EnSF or LETKF to update the posterior filtering density after simulating the Lorenz-96 model $10$ time steps.

% \vspace{0.5em}

\subsubsection{Hyper-parameter fine tuning}\label{sec:fine}
% \noindent \textbf{Fine-tuning}
% \vspace{0.5em}

\begin{figure}[h!]
\centering
\includegraphics[width=0.95\textwidth]{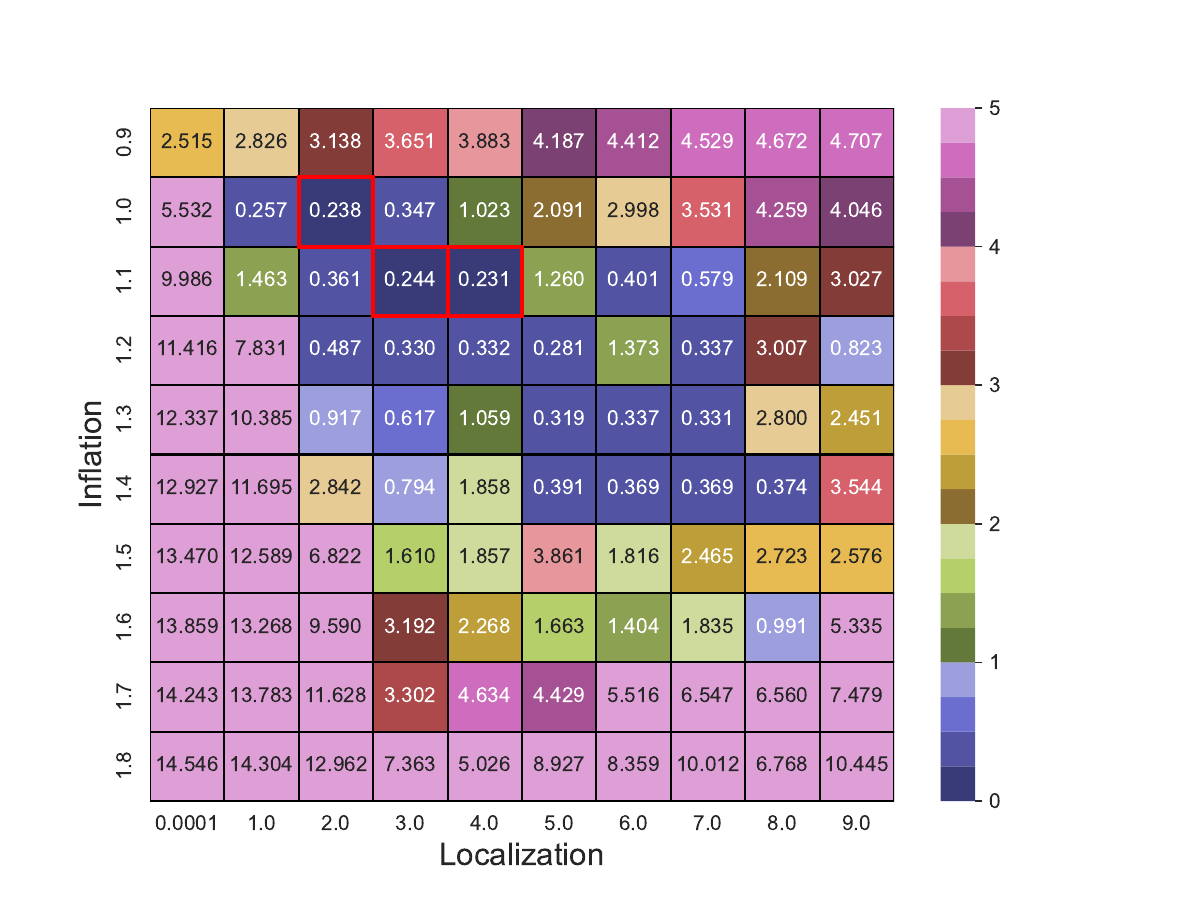}
\vspace{-1em}
\caption{LETKF's fine-tuning chart where the RMSE is averaged on all data assimilation times with 10 repetitions. The highlighted cells are the best three parameter combinations selected for LETKF. }\label{LETKF_Tune_All}
\end{figure}

\begin{figure}[h!]
\centering
\includegraphics[width=0.95\textwidth]{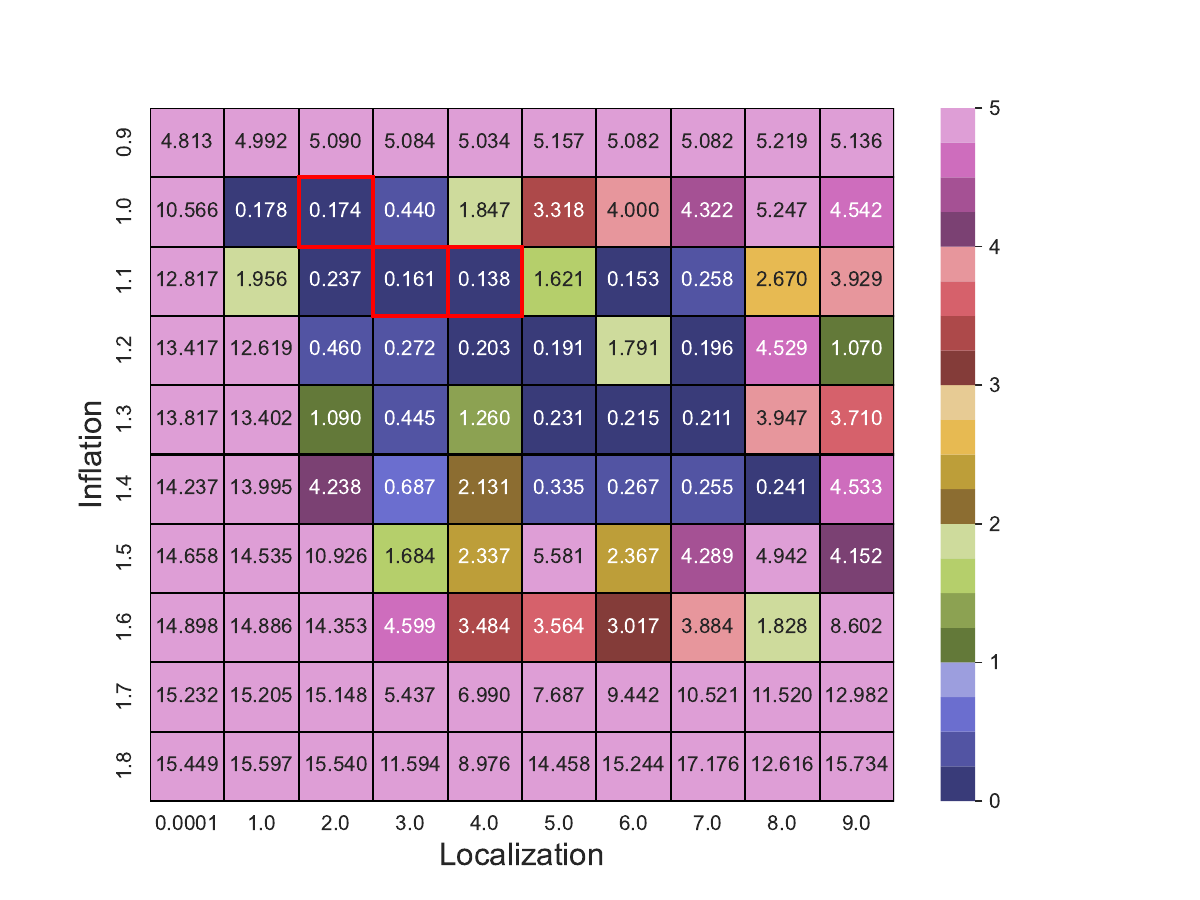}
\vspace{-1em}
\caption{LETKF's fine-tuning chart where the RMSE is averaged on the last 50 data assimilation times with 10 repetitions.  The highlighted cells are the best three parameter combinations selected for LETKF. }\label{LETKF_Tune_last}
\end{figure}

An important feature of LETKF is that it utilizes an inflation factor and a localization factor to fine-tune the behavior of the ensemble Kalman filter so that the fine-tuned LETKF can fit a specific optimal filtering problem. Therefore, before we conduct the comparison experiments, we first fine-tune LETKF. According to the computational cost shown in Figure \ref{Efficiency}, it takes around 300 seconds for LETKF to perform one filtering step in tracking the 1,000,000-dimensional Lorenz-96 model. This means finishing the fine-tuning chart for LETKF in Figure \ref{LETKF_Tune_All} and \ref{LETKF_Tune_last} for the one million dimensional cases will cost about 
520 days using a single RTX 3070 GPU, which is not practical. Therefore, we perform LETKF fine-tuning in the 
in the $100$ dimensional space ($d = 100$), which costs around 2 hours to generate the fine-tuning charts. Additionally, fine-tuning in 100-dimensional space and testing in 1,000,000-dimensional space will demonstrate the transferability of LETKF and EnSF.

Specifically, we let the inflation factor vary from $0.9 - 1.8$, and the localization factor is tested from $0.0001 - 9$ \footnote{The corresponding testing ranges for inflation and localization are already optimized based on our experience. In practice, one may need to test the inflation factor and the localization factor in much larger ranges}, and the ensemble size that we picked for LETKF is $20$. Then, we solve the corresponding Lorenz-96 tracking problem repeatedly $10$ times, and the overall RMSEs averaged on all data assimilation times are presented in Figure \ref{LETKF_Tune_All}. We also use a colorbar to visually represent the various RMSEs and provide an intuitive understanding of the fine-tuned results. In Figure \ref{LETKF_Tune_last}, we present the average RMSEs (over $10$ repetitions) on the last $50$ data assimilation times, which indicates the converged performance of LETKF in the tuning procedure. Based on Figure \ref{LETKF_Tune_All}, we choose the best three parameter combinations for LETKF:
\begin{itemize}
    \item LETKF (No.1): Inflation=1.1, localization=4;
    \item LETKF (No.2): Inflation=1.0, localization=2;
    \item LETKF (No.3): Inflation=1.1; localization=3.
\end{itemize}
The selected LETKF parameters are highlighted in both Figure \ref{LETKF_Tune_All} and Figure \ref{LETKF_Tune_last}, and will be used for further comparisons with EnSF.

% From these figures, we can see that the optimal inflation factor is $1.1$ and the optimal localization factor is $4$. {\color{red} add 3 parameter values}

\begin{figure}[h!]
\centering
\includegraphics[width=0.95\textwidth]{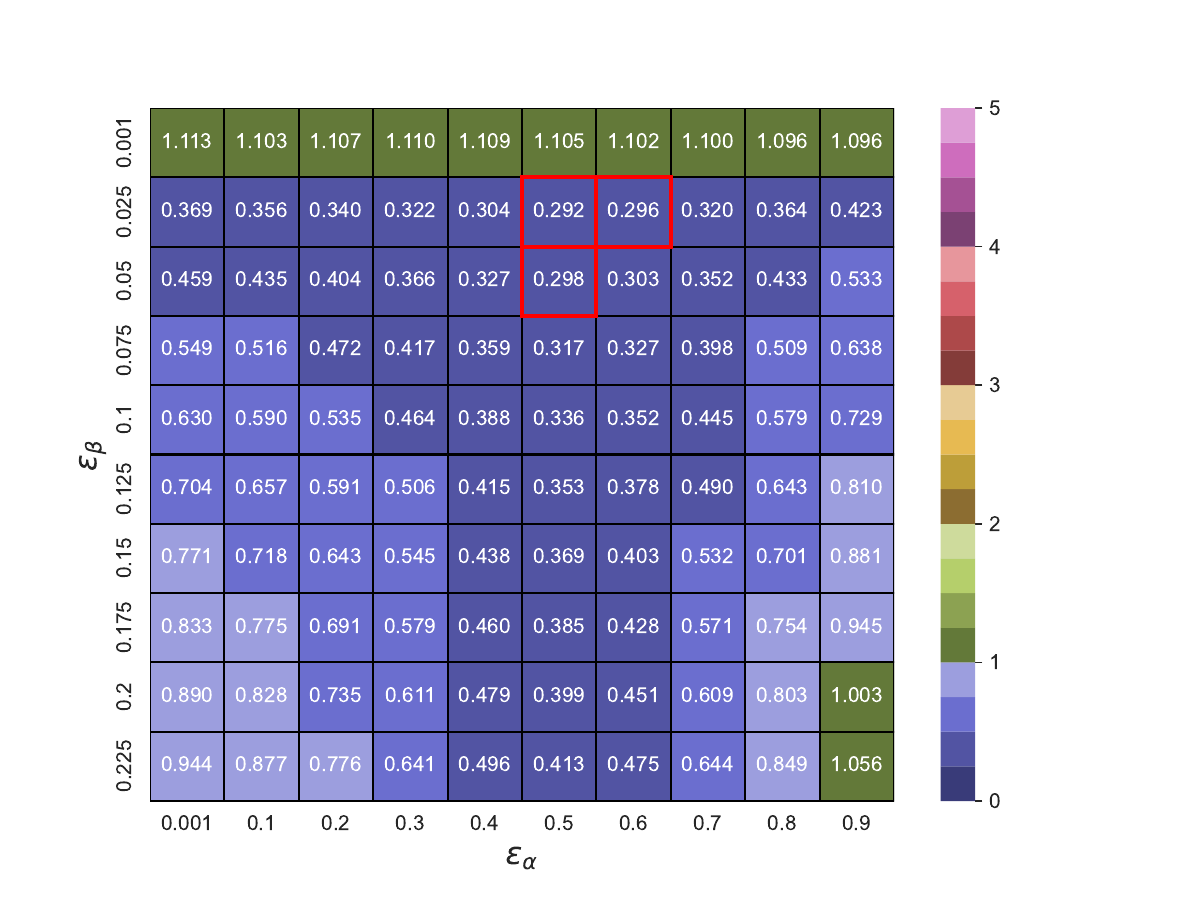}
\vspace{-1em}
\caption{EnSF's fine-tuning chart where the RMSE is averaged on all data assimilation times with 10 repetitions.  The highlighted cells are the best three parameter combinations selected for EnSF.  Compared to LETKF, EnSF's performance is much more stable with respect to small changes of the hyper-parameters.}\label{EnSF_Tune_All}
\end{figure}
\begin{figure}[h!]
\centering
\includegraphics[width=0.95\textwidth]{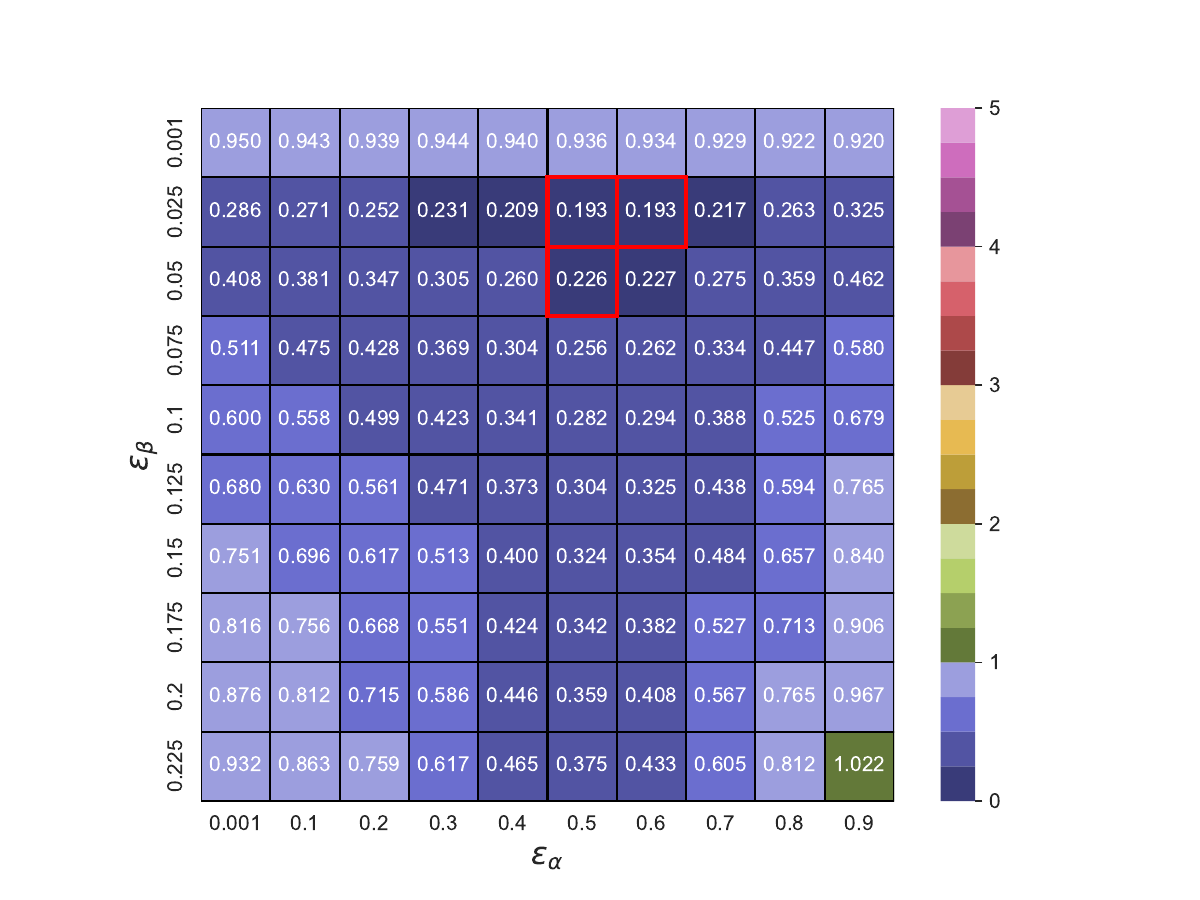}
\vspace{-1em}
\caption{EnSF's fine-tuning chart where the RMSE is averaged on the last 50 data assimilation times with 10 repetitions. The highlighted cells are the best three parameter combinations selected for EnSF. Compared to LETKF, EnSF's performance is much more stable with respect to small changes of the hyper-parameters.}\label{EnSF_Tune_last}
\end{figure}

To compare with LETKF, we also fine-tune EnSF's hyperparameters, which are introduced in Eq.~\eqref{eq:hyper}. In Figures \ref{EnSF_Tune_All} and \ref{EnSF_Tune_last}, we present the fine-tune charts for EnSF under the same setup as LETKF, with the RMSEs presented in each block marked by the same colorbar as used for LETKF fine-tune charts. The best three parameter combinations for EnSF are highlighted in both figures and will be used for comparison. They are as follows: 
\begin{itemize}
    \item EnSF (No.1): $\epsilon_\alpha=0.5$, $\epsilon_\beta=0.025$;
    \item EnSF (No.2): $\epsilon_\alpha=0.6$, $\epsilon_\beta=0.025$;
    \item EnSF (No.3): $\epsilon_\alpha=0.5$, $\epsilon_\beta=0.05$.
\end{itemize}
We can see from the fine-tune charts for LETKF and EnSF that both methods achieve comparable accuracy with their optimal hyperparameters. However, EnSF is less sensitive to these hyperparameters with a wide range of configurations yielding good performance. On the other hand, the performance of LETKF varies dramatically, which indicates that it is very sensitive to the choice of inflation factor and localization factor. Such sensitivity may cause additional difficulty when attempting to fine-tune LETKF in more complex problems.

\subsubsection{Efficiency comparison between EnSF and LETKF}\label{sec:efficincy}
% \vspace{1em}
% \noindent \textbf{Efficiency comparison}
% \vspace{1em}

To proceed, we first carry out an efficiency comparison between EnSF and then show the performance comparison in solving the one million dimensional problem.
\begin{figure}[h]
\centering
\includegraphics[width=0.75\textwidth]{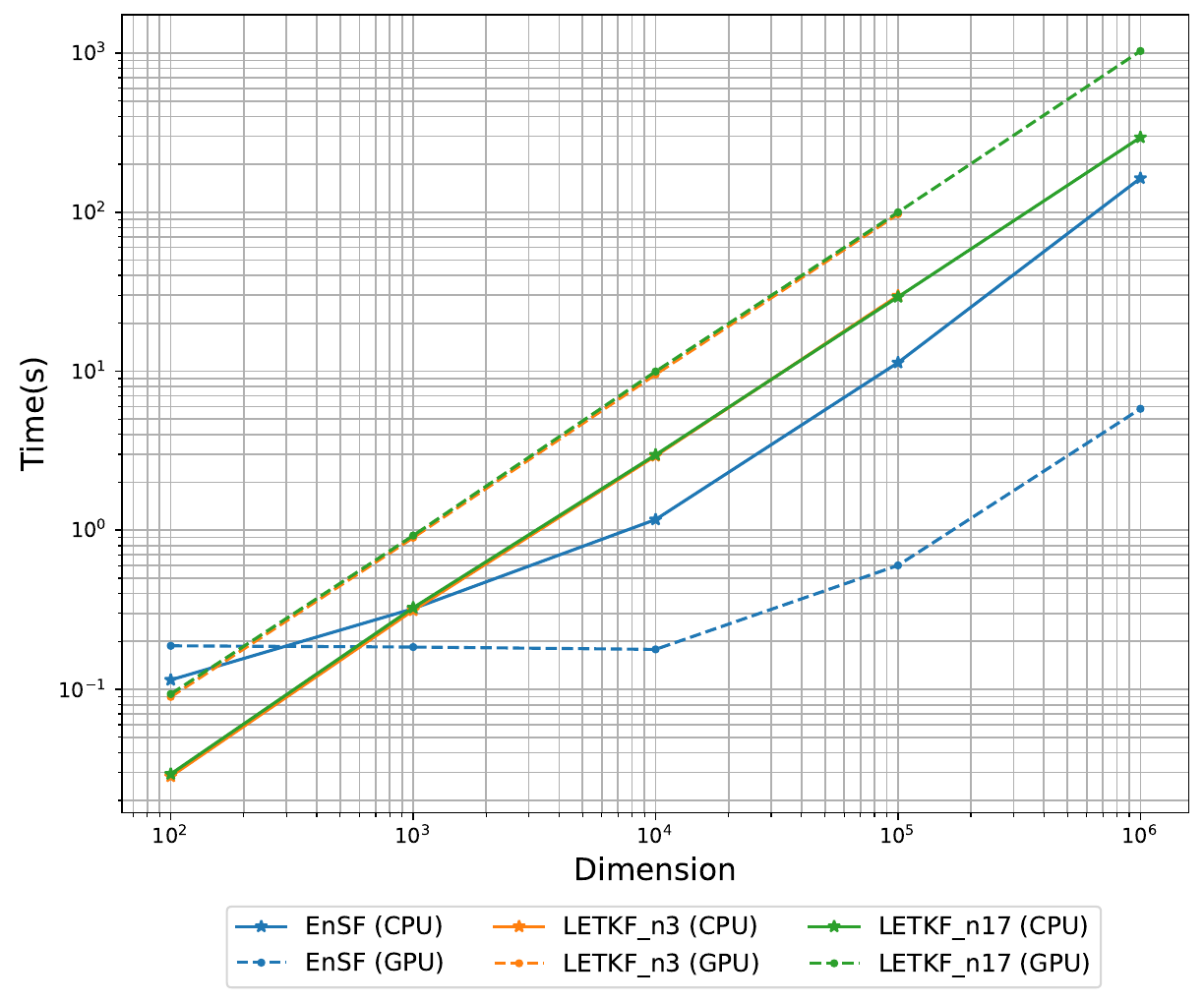}
% \vspace{-1em}
\caption{One step data assimilation computational cost with ensemble size $=20$. EnSF with GPU implementation is much more efficient than LETKF, and EnSF is more suitable for modern GPU machines.}\label{Efficiency}
\end{figure}
In Figure \ref{Efficiency}, we showcase the computational cost for implementing \textit{one data assimilation step} of EnSF and LETKF in solving problems ranging from $100$ dimensions up to $1,000,000$ dimensions. The ensemble size chosen for both methods is 20. The CPU used is a 6-core AMD Ryzen™ 5 5600X, and the GPU used is an NVIDIA RTX 3070. Both EnSF and LETKF are implemented on CPU and GPU. The LETKF is tested with neighbor sizes of 3 (LETKF\_n3) and 17 (LETKF\_n17). The neighbor size refers to the number of state variables retained for calculating the localized covariance, which depends on the value of the localization parameter. % LETKF is implemented with a localization factor of 1 (LETKF\_n1) and 8 (LETKF\_n8) with an inflation factor chosen as $1$. 
The one-step data assimilation computing time is calculated as the average of $20$ repetitions. From this efficiency figure, we can see that EnSF is much more efficient than LETKF. From the $100$ dimension to the $10,000$ dimension, the main computational cost for EnSF is essentially the background computation, and it's only $0.17$ second per step on average. For the $1,000,000$ dimensional problem, the average computational cost for EnSF is only approximately $5$ seconds per step. On the other hand, the computational cost of LETKF is approximately $300$ seconds per step when implemented with 20 Kalman filter samples. Although a 6-core CPU could potentially run LETKF faster, the total computational cost of LETKF is still much higher than that of EnSF, and LETKF is not suitable for modern GPU machines, which makes it difficult to further scale LETKF algorithms in practical implementations. Due to the extremely high computational cost of LETKF in solving the 1,000,000-dimensional problem, it is not feasible to fine-tune LETKF in the one-million-dimensional space. Therefore, we utilize the optimal hyperparameters for both methods that we obtained in the 100-dimensional space to run the 1,000,000-dimensional problem.

\subsubsection{Experimental setting 1: baseline test}\label{sec:base}
% \vspace{1em}
% \noindent \textbf{Experimental setting 1: baseline test}
% \vspace{1em}

\begin{figure}[h!]
\centering
\includegraphics[width=0.75\textwidth]{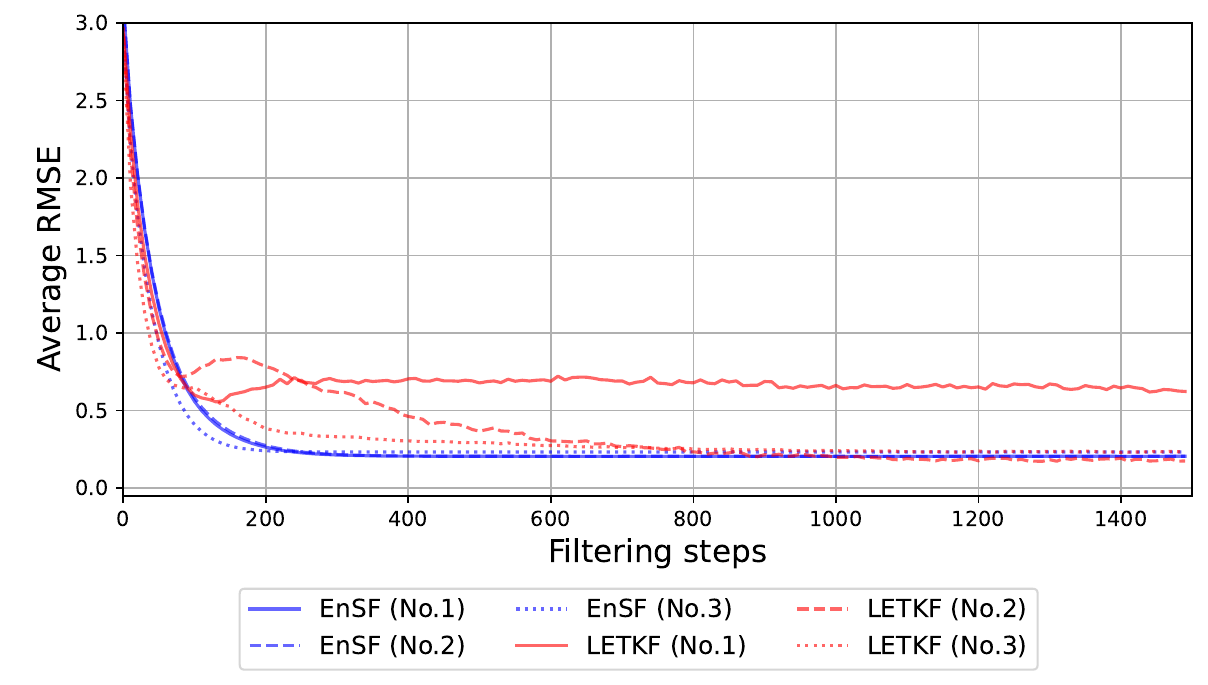}
% \vspace{-0.4cm}
\caption{RMSEs comparison between EnSF and LETKF at data assimilation times. RMSEs are calculated by repeating the same test 10 times with different random initial conditions, and we average the estimation errors over all 1,000,000 directions and 10 repetitions. No.1, No.2, and No.3 in the legend correspond to the first, second, and third-best hyperparameters, respectively. We observe that EnSF performs consistently well with the top three sets of hyperparameters, while the accuracy of LETKF varies.}\label{1MD_baseline}
\end{figure}

We first conduct a baseline comparison using the same problem setting that was used to fine-tune LETKF and EnSF, except that the dimension of the Lorenz-96 model is now $d=1,000,000$. The comparison of root mean square errors (RMSEs) at data assimilation times is presented in Figure \ref{1MD_baseline}, where we have selected the top three sets of hyperparameters for each method, and the RMSEs are plotted with respect to time. In our numerical experiments, RMSEs are calculated by repeating the same test 10 times with different random initial conditions, and we average the estimation errors over all 1,000,000 directions and 10 repetitions. We can see that both EnSF and LETKF provide good accuracy for tracking the target Lorenz-96 state. EnSF performs consistently well with three different sets of hyperparameters, while the accuracy of LETKF varies.

\begin{figure}[h!]
\centering
\includegraphics[width=0.75\textwidth]{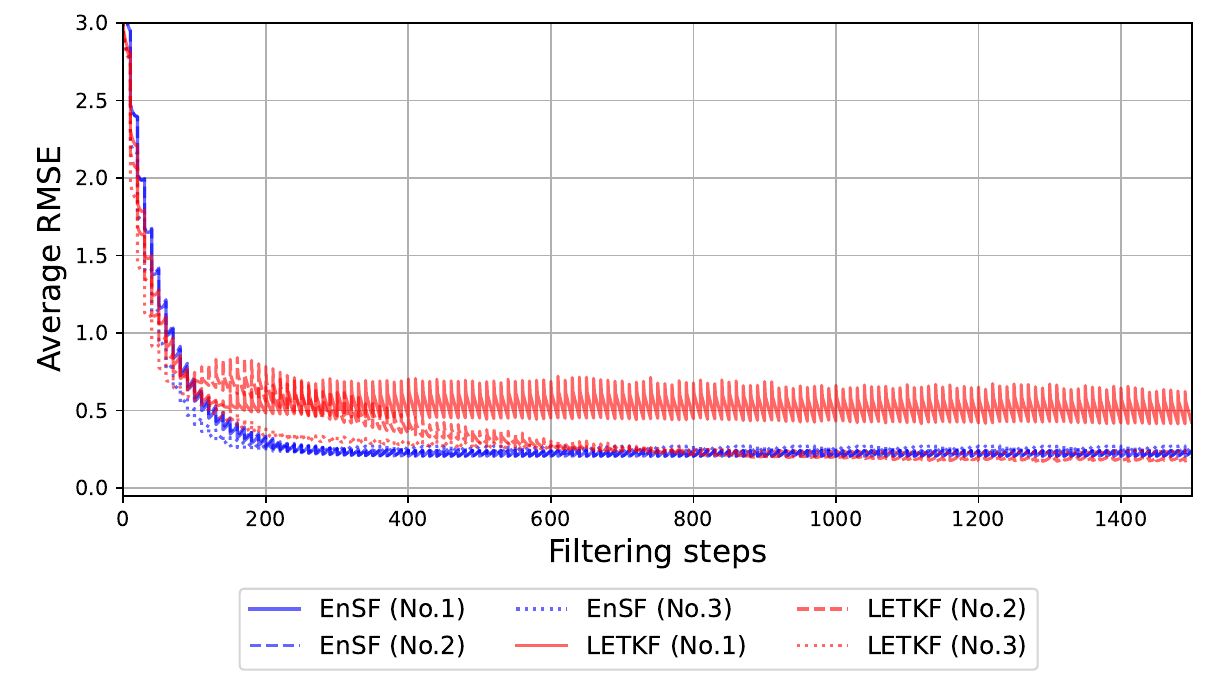}
% \vspace{-0.4cm}
\caption{RMSEs comparison between EnSF and LETKF at every time step -- including prediction-only steps and data assimilation steps. Compared to Figure \ref{1MD_baseline}, we observe fluctuations in estimation errors for both methods.}\label{1MD_baseline_everyone}
\end{figure}

Figure \ref{1MD_baseline_everyone} shows the comparison of RMSEs at every time step -- including prediction-only steps and data assimilation steps where prediction and Bayesian updates are performed.
From this figure, we can see fluctuations in estimation errors for both methods. When providing good accuracy, the difference between prediction-only errors and data assimilation errors is small. However, for one LETKF test, which provided the least accurate result, there's a larger variance between prediction-only steps and data assimilation steps. This partially explains why the hyperparameter choice ``No.1'' of LETKF did not work as well as the other two.

\subsubsection{Experimental setting 2: reduced observation noise test}\label{sec:noise}
\begin{figure}[h!]
\centering
\includegraphics[width=0.75\textwidth]{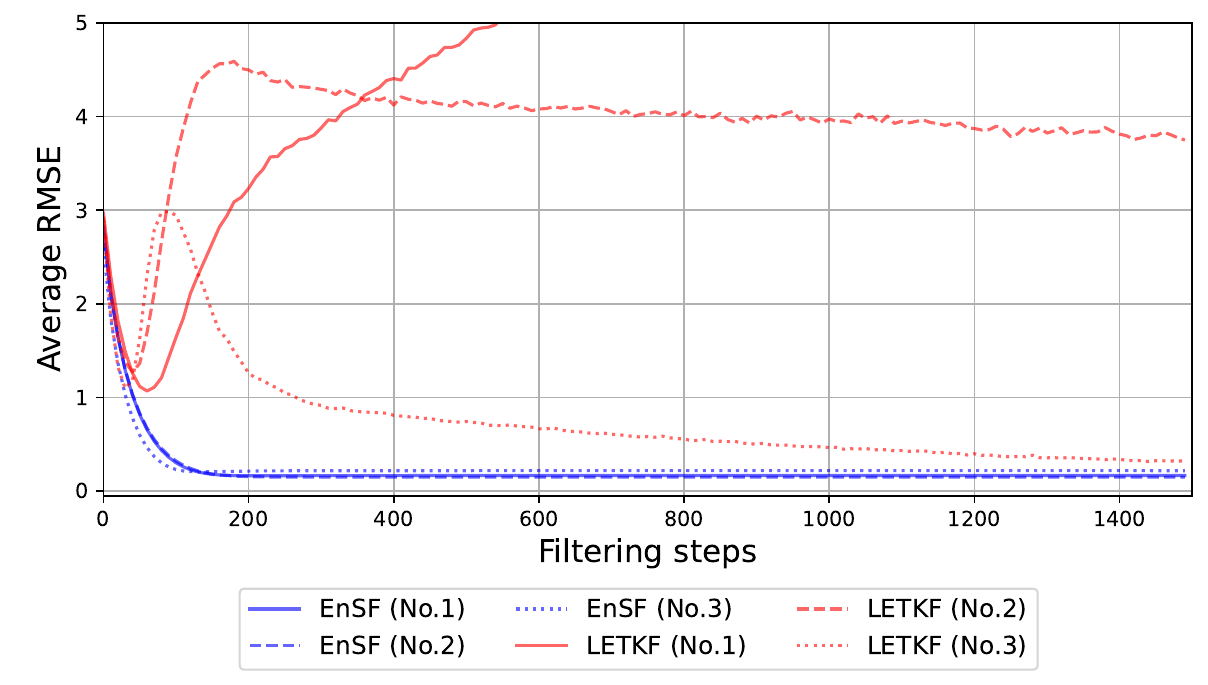}
% \vspace{-0.4cm}
\caption{RMSEs comparison between EnSF and LETKF at data assimilation time steps with smaller observational noise $\varepsilon_t \sim(\mathbf{0}, 0.03^2\mathbf{I}_{d})$. 
We observe that although the observational data are more accurate, two of the top choices of hyperparameters for LETKF diverge. In comparison, EnSF continues to provide very accurate estimates for the target state.}\label{1MD_smaller}
\end{figure}

\begin{figure}[h!]
\centering
\includegraphics[width=0.75\textwidth]{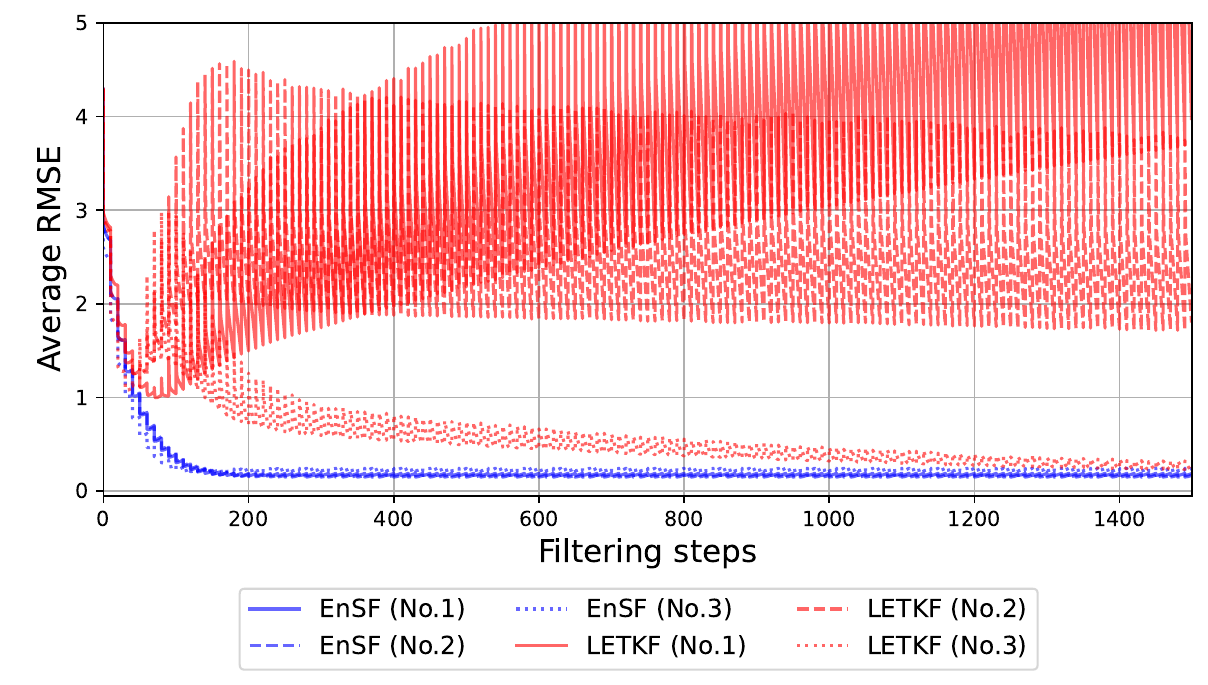}
% \vspace{-0.4cm}
\caption{RMSEs comparison between EnSF and LETKF at every time step with smaller observation noise $\varepsilon_t \sim(\mathbf{0}, 0.03^2\mathbf{I}_{d})$. 
}\label{1MD_smaller_everyone}
\end{figure}

Next, we modify the problem setting by reducing the observational noise from $\varepsilon_t \sim(\mathbf{0}, 0.05^2\mathbf{I}_{d})$ to $\varepsilon \sim(\mathbf{0}, 0.03^2\mathbf{I}_{d})$ and conduct the same comparison experiment. The corresponding RMSEs at data assimilation time steps are presented in Figure \ref{1MD_smaller}, and the RMSEs at all time steps are presented in Figure \ref{1MD_smaller_everyone}.
\begin{figure}[h!]
\centering
\includegraphics[width=0.75\textwidth]{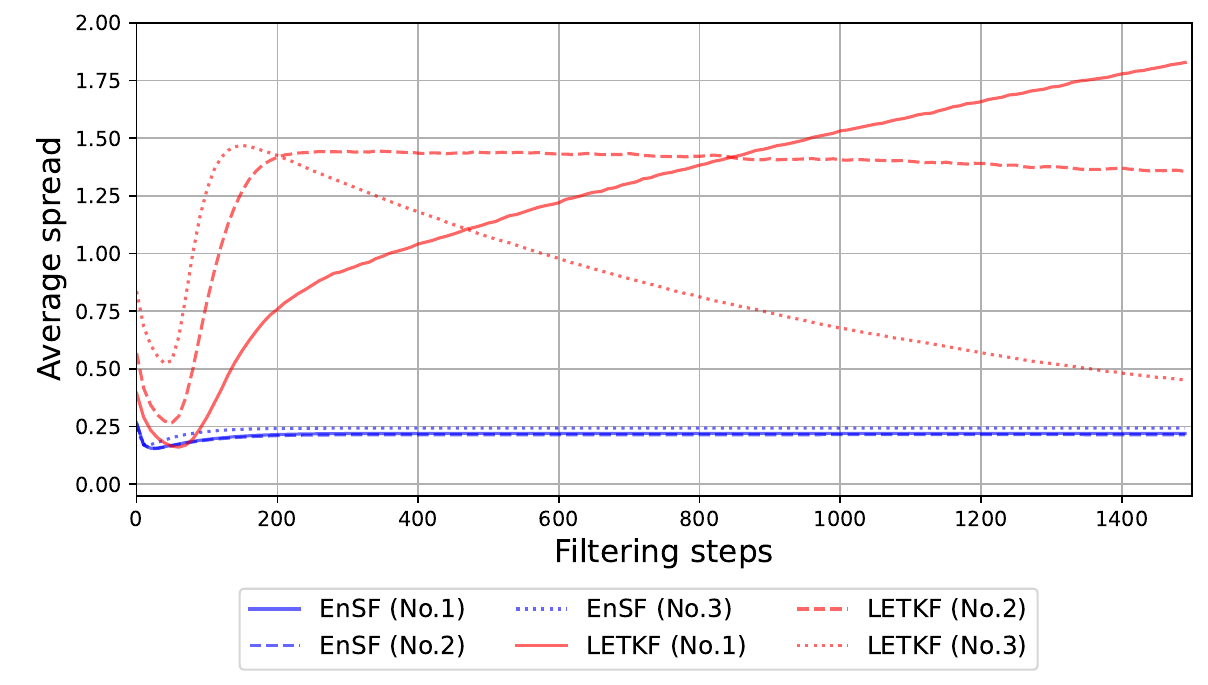}
% \vspace{-0.4cm}
\caption{Average ensemble spread in the smaller observation noise test. While a larger ensemble spread allows a filtering method to better cover the true signal, overly wide spread state samples provide less useful information about the true target state. }\label{1MD_smaller_spread}
\end{figure}

Although the observational data are more accurate in this experiment, two of the top choices of hyperparameters for LETKF diverge, and the hyperparameter that provides the best result in this experiment is actually the third-best choice from the fine-tune chart. This result shows that while the fine-tuned LETKF provides higher accuracy, it is very sensitive to the problem setting. Even slight modifications to the problem may cause severe divergence in LETKF, with no indication beforehand which set of hyperparameters will fail. In addition, the all-time RMSEs presented in Figure \ref{1MD_smaller_everyone} verify that LETKF failed with hyperparameters No.1 and No.2. In comparison, EnSF continues to provide very accurate estimates for the target state for the top-three choices of its hyper-parameters.

To better illustrate the performance of EnSF and LETKF, we plot the average ensemble spread in Figure \ref{1MD_smaller_spread}. The ensemble spread is the square root of the average ensemble variance, i.e., $\sqrt{|| Var(\bx_{ensemble}) ||_1 / d}$, where the variance is calculated for each dimension of the ensemble. While a larger ensemble spread allows a filtering method to better cover the true signal, overly wide spread state samples provide less useful information about the true target state, which makes the predicted state less reliable. This issue is also reflected in the all-time RMSEs shown in Figure \ref{1MD_smaller_everyone}, where the hyperparameters causing widely spread LETKF samples correspond to large fluctuations in the RMSEs. On the other hand, EnSF maintains very stable ensemble spread sizes, and the best-performing LETKF shows a converging ensemble spread.

% \begin{figure}[H]
% \centering
% \includegraphics[width=0.7\textwidth]{figs_new/1MD_rep/crps_prior_prob_1.png}
% \caption{Average CRPS (smaller obs noise: 0.03)}\label{1MD_smaller_CRPS_prior}
% \end{figure}

% \begin{figure}[H]
% \centering
% \includegraphics[width=0.7\textwidth]{figs_new/1MD_rep/crps_post_prob_1.png}
% \caption{Average CRPS (smaller obs noise: 0.03)}\label{1MD_smaller_CRPS_posterior}
% \end{figure}

\subsubsection{Experimental setting 3: incomplete knowledge of the model error}
% \vspace{1em}
% \noindent \textbf{Experimental setting 3: incomplete knowledge test}
% \vspace{1em}
%More specifically, for a given relative shock size $s$, i.e., $5\%$, $20\%$, and $50\%$, the shock is added as 
%\begin{equation}
%    \bx = \bx + s |\bx| \odot \bz
%\end{equation}
%where $\bz \mN(\bzero, \bI)$ and $\odot$ is element-wise product. In other words, the size of the shock is proportional to the size of the state value. 

In the last experimental setting, we address a more challenging but realistic scenario involving \textit{an imperfect model due to incomplete knowledge}. In this scenario, we assume that the state model may not fully reflect the true state propagation, and we model this unknown portion as a mixture of three levels of independent Gaussian-type random shocks. On the evolution of the true state trajectory, we introduce independent shocks with probabilities of $2\%$, $1\%$, and $0.5\%$, with corresponding shock sizes of $5\%$, $20\%$, and $50\%$ relative to the current state magnitude of the Lorenz-96 model, respectively. For example, when a size $50\%$ shock happens, every component of the true state $\bx_t^i$ is perturbed by an additive term $0.5 Z_i |\bx_t^i|$, where $\bx_t^i$ is the i-th dimension of the true state and $Z_i$ are i.i.d. standard Gaussian noise.

This problem setting mimics a situation where there is a small chance (2\%) that the model is inaccurate with a 5\% error. There's an even smaller chance (1\%) that the model error is larger, at 20\% level, and a very small chance (0.5\%) that a large-scale unexpected error occurs in the model. In practical applications, this scenario is quite common due to the limited knowledge we have about the real world. For example, in weather forecasting, this variety of unknown model errors is used to simulate the effects of flow-dependent model uncertainties (see discussions in \cite{held_et_al_1995}).

\begin{figure}[h!]
\centering
\includegraphics[width=0.75\textwidth]{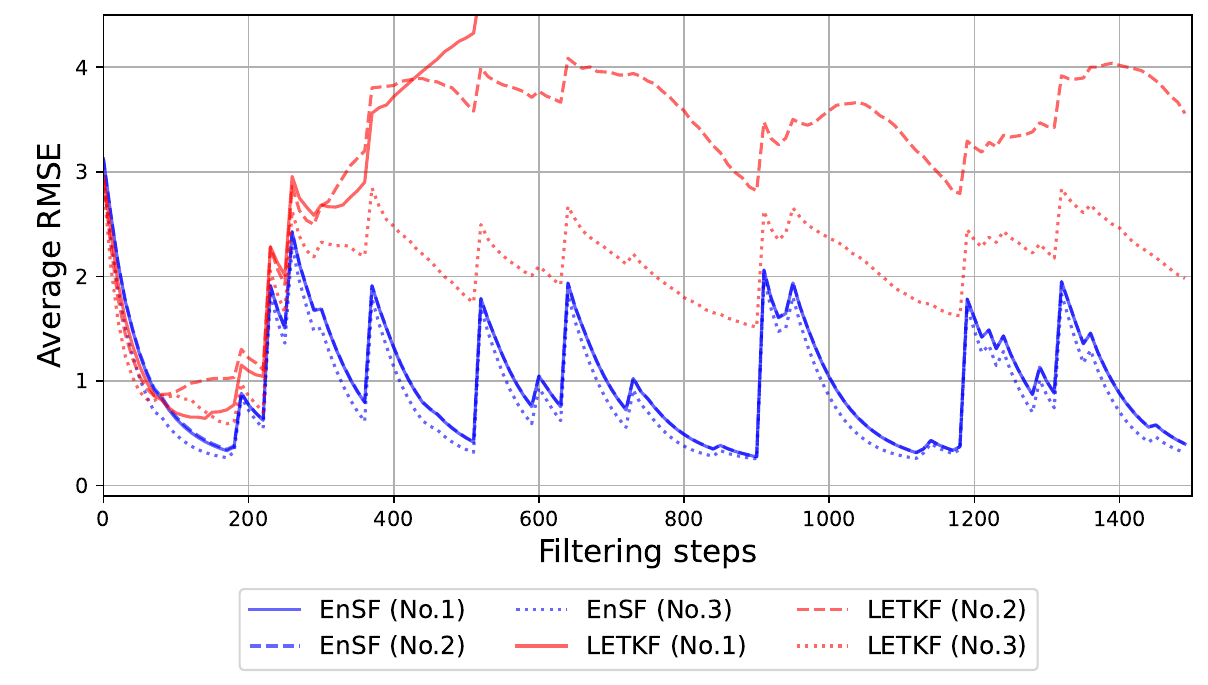}
% \vspace{-0.4cm}
\caption{RMSEs comparison between EnSF and LETKF in the incomplete knowledge experiment, where the unknown model error is injected into the state equation as random shocks. We observe that EnSF can quickly recover from unexpected shocks, but  LETKF either
diverges or struggles to recover from the shock.}\label{1MD_shocks}
\end{figure}

\begin{figure}[h!]
\centering
\includegraphics[width=0.95\textwidth]{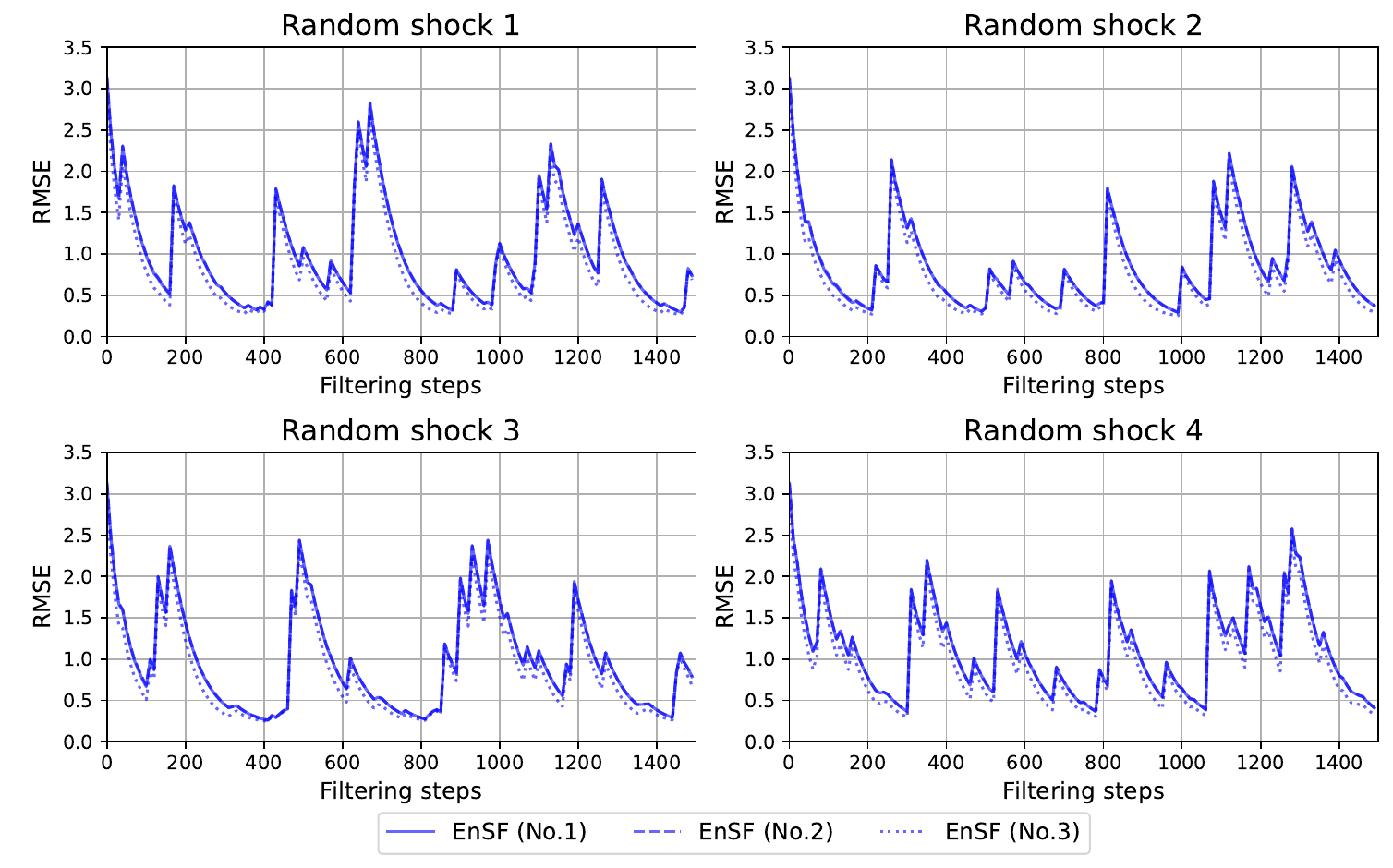}
% \vspace{-0.4cm}
\caption{Repeated experiments of EnSF in the incomplete knowledge scenarios, each subfigure shows the RMSE of EnSF for a different occurrence of random shocks. We observe that EnSF performs stably with different random shock patterns.}\label{1MD_shocks_repeat}
\end{figure}

In Figure \ref{1MD_shocks}, we compare the RMSEs, which are calculated by averaging the estimation errors over all 1,000,000 directions, of EnSF with LETKF in the imperfect model scenario at data assimilation time steps. The observational noise is kept at $\varepsilon_t \sim(\mathbf{0}, 0.05^2\mathbf{I}_{d})$, which is the setting used for fine-tuning. The figure also shows the time instants when the unexpected shocks occur during the data assimilation period. We can see from Figure \ref{1MD_shocks} that all three settings of EnSF can quickly recover from unexpected shocks. However, LETKF either diverges or struggles to recover from the shocks. Unlike the previous test, where fine-tuning is possible for a smaller observational noise value, the unexpected shocks in this experiment are caused by the ``unknown'' portion of the state model. Since we lack information about these unknown shocks, we cannot fine-tune either EnSF or LETKF.  
To further validate the reliable performance of EnSF, we repeat the above experiments four times with different occurrences of random shocks and show the corresponding tracking RMSEs of EnSF in Figure \ref{1MD_shocks_repeat}. From this figure, we can see that EnSF consistently generates low errors and quickly recovers from unexpected shocks.

\begin{figure}[h!]
\centering
\includegraphics[width=0.75\textwidth]{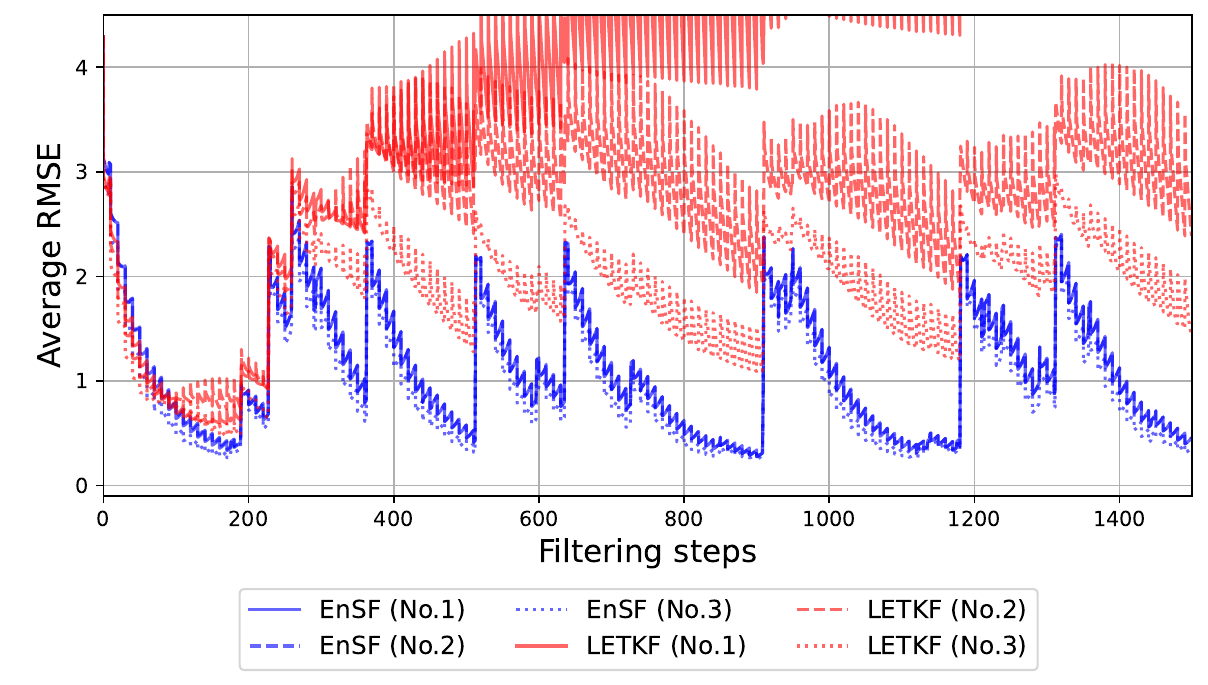}
% \vspace{-0.4cm}
\caption{Comparison between EnSF and LETKF at every time step in the incomplete knowledge experiment, where the unknown model error is injected into the state equation as random shocks. We observe that LETKF has highly fluctuating estimation errors, but EnSF's error fluctuation between two filtering steps is much smaller. }\label{1MD_shocks_everyone}
\end{figure}

\begin{figure}[h!]
\centering
\includegraphics[width=0.75\textwidth]{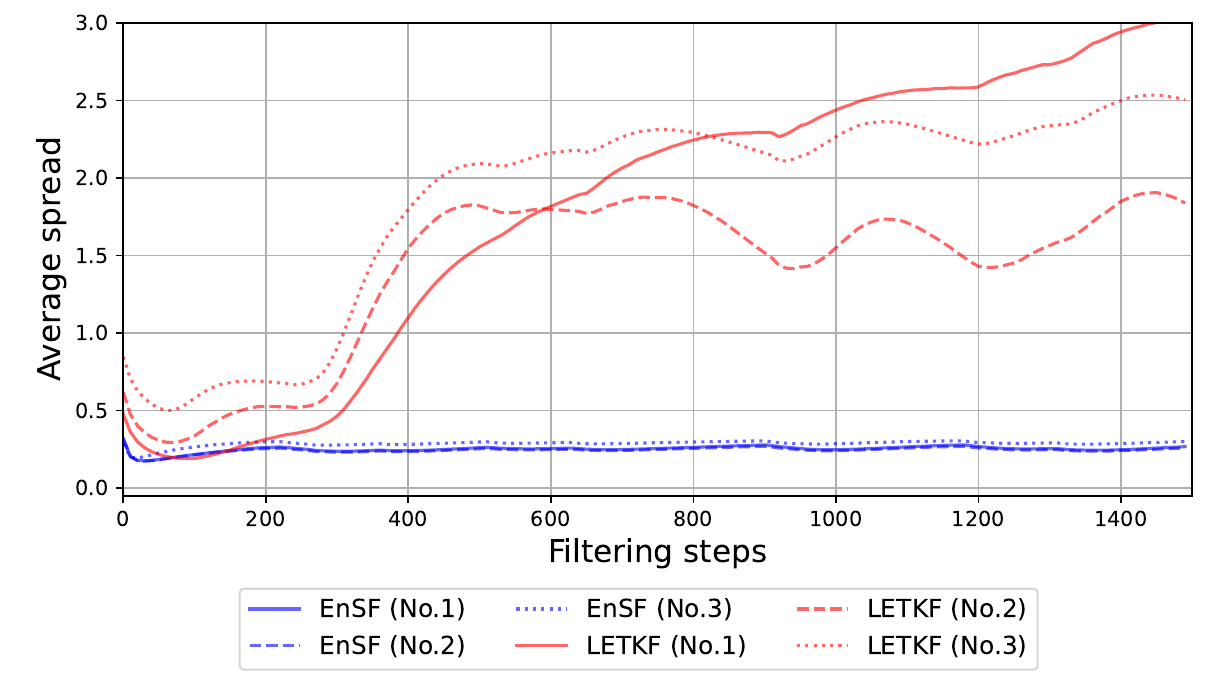}
% \vspace{-0.4cm}
\caption{Average ensemble spread in the incomplete knowledge experiment, where the unknown model error is injected into the state equation as random shocks. We observe that the ensemble spreads of LETKF are generally wide or even divergent.}\label{1MD_shocks_spread}
\end{figure}

In Figure \ref{1MD_shocks_everyone}, we present the RMSE comparison at every time step, and in Figure \ref{1MD_shocks_spread} we show the comparison of average ensemble spreads in this incomplete knowledge experiment. From these figures, we can see that LETKF has highly fluctuating estimation errors, and its average ensemble spreads are generally wide or even divergent. These two pieces of evidence indicate that LETKF is not stable enough to handle unknown model errors due to incomplete knowledge or information. 
To further validate the reliable performance of EnSF, we repeat the above experiments four times with different realizations of random shocks and show the corresponding tracking RMSEs of EnSF in Figure \ref{1MD_shocks_repeat}. From this figure, we can see that EnSF constantly generates low errors, and it always recovers from unexpected shocks quickly.
%
% \begin{figure}[h!]
% \centering
% \includegraphics[width=0.45\textwidth]{figs_new/1MD/EnSF_shock_1.png}
% \includegraphics[width=0.45\textwidth]{figs_new/1MD/EnSF_shock_2.png}\\
% \includegraphics[width=0.45\textwidth]{figs_new/1MD/EnSF_shock_3.png}
% \includegraphics[width=0.45\textwidth]{figs_new/1MD/EnSF_shock_4.png}
% \vspace{-0.4cm}
% \caption{Repeated experiments of EnSF in the incomplete knowledge scenarios, each subfigure shows the RMSE of EnSF for a different occurrence of random shocks. We observe that EnSF performs stably with different random shock patterns.}\label{1MD_shocks_repeat}
% \end{figure}

\begin{figure}[h!]
\centering
\includegraphics[width=0.75\textwidth]{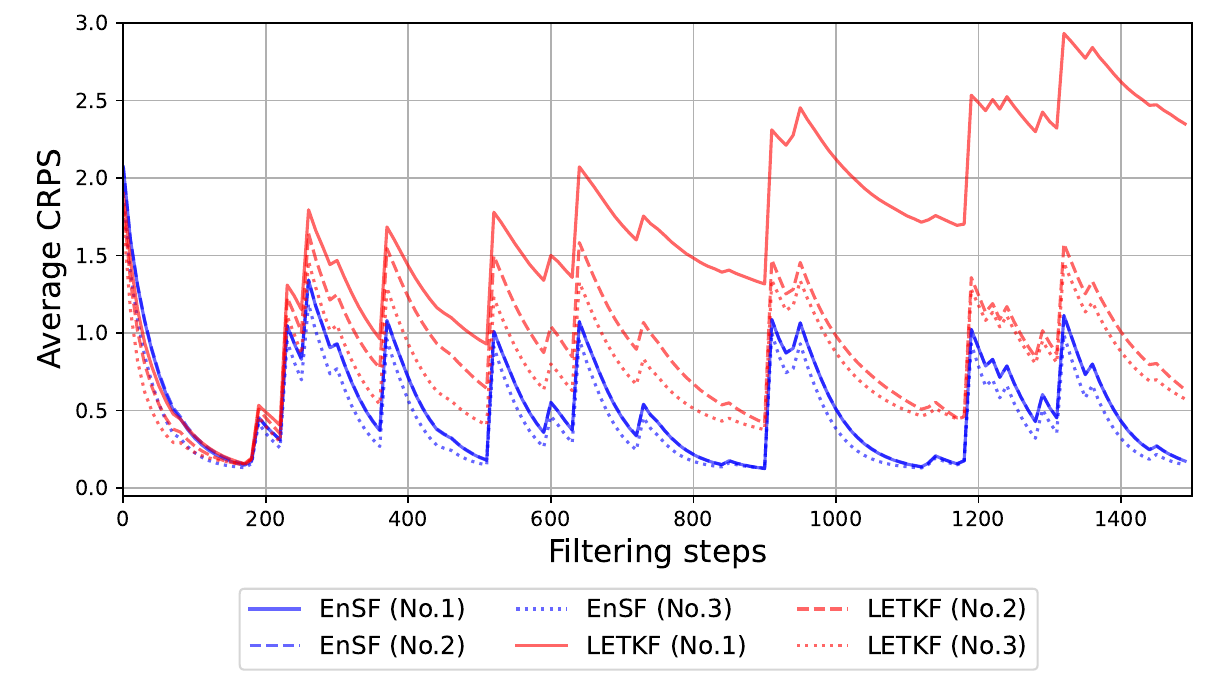}
% \vspace{-0.4cm}
\caption{CRPS comparison for posterior filtering densities. Lower CRPS values indicate that the distribution is well-aligned with the true state. We observe that EnSF stably outperforms LETKF in this experiment.}\label{1MD_shocks_CRPS_posterior}
\end{figure}

In this experimental setting, we introduce an extra metric to evaluate the performance of EnSF and LETKF, namely the continuous ranked probability score (CRPS). Specifically, at a given time step, for a marginal dimension $i$, let $F^{i}_{ensmeble}(z)$ and $F^{i}_{true}(z)$ be the empirical cumulative distribution functions of the ensemble and the true state, respectively, where $F^{i}_{true}(z) = 1_{z>\bx_{true}^{i}} (z)$. Then, we let $CRPS^i := \int (F^{i}_{ensmeble}(z) - F^{i}_{true}(z))^2 dz$, and we average the CRPS over all $1,000,000$ dimensions. The CRPS measures how well the ensemble distribution matches with the true state. Lower CRPS values indicate that the ensemble distribution is well-aligned with the true state with a small uncertainty, while higher CRPS values indicate otherwise. In this experiment, we plot the CRPS comparison between EnSF and LETKF at data assimilation steps in Figure \ref{1MD_shocks_CRPS_posterior}. From this figure, we can see that EnSF outperforms LETKF in this CRPS comparison.

\section{Conclusion}\label{sec:con}
We propose the EnSF method to solve very high-dimensional nonlinear filtering problems. The avoidance of training neural networks to approximate the score function makes it computationally feasible for EnSF to efficiently solve the 1,000,000-dimensional Lorenz-96 problem. We observe in the numerical experiments that one million dimensions is definitely not the upper limit of EnSF's capability, especially with the help of modern high-performance computing. Besides trying high-dimensional cases, there are several key aspects of EnSF that can be improved in the future. 
First, we will investigate how fast the number of samples, i.e., $J$ in Algorithm 1, needs to grow with the dimensionality to ensure robust performance. Second, we will expand the capability of the current EnSF to handle partial observations, i.e., only a subset of the state variables are involved in the observation process, which is critical to real-world data assimilation problems. In fact, Figure \ref{obsshow} shows that the $arctan()$ observation function can be reviewed as a partial observation in the sensing that there is no observational information when the state is outside 
$[-\pi/2,\pi/2]$. Third, the current definition of the weight function $h(\tau)$ in Eq.~\eqref{eq:d} for incorporating the likelihood into the score function is empirical. The current choice of $h(\tau)$ may introduce a bias into the posterior state estimation. We will investigate whether there is an optimal weight function to gradually incorporate the likelihood information into the backward SDEs. Fourth, the efficiency of backward sampling can also be improved by incorporating advanced stable time-stepping schemes, e.g., the exponential integrator, to significantly reduce the number of time steps in the discretization of the backward process in the diffusion model. Fifth, we will test the performance of EnSF for real-world models, e.g., the IFS model developed by ECMWF, and the existing AI-based weather models, e.g., FourCastNet, GraphCast. 

\section*{Acknowledgement}

This material is based upon work supported by the U.S. Department of Energy, Office of Science, Office of Advanced Scientific Computing Research, Applied Mathematics program under the contract ERKJ387 at the Oak Ridge National Laboratory, which is operated by UT-Battelle, LLC, for the U.S. Department of Energy under Contract DE-AC05-00OR22725. The first author (FB) would also like to acknowledge the support from U.S. National Science Foundation through project DMS-2142672 and the support from the U.S. Department of Energy, Office of Science, Office of Advanced Scientific Computing Research, Applied Mathematics program under Grant DE-SC0022297.

\bibliographystyle{siam}
\bibliography{library,escape_ref,nfref,Reference}

\end{document}